\documentclass{article}

\usepackage[main, final]{neurips_2026}

\usepackage[utf8]{inputenc} 
\usepackage[T1]{fontenc}    
\usepackage{hyperref}       
\usepackage{url}            
\usepackage{booktabs}       
\usepackage{amsfonts}       
\usepackage{nicefrac}       
\usepackage{microtype}      
\usepackage{xcolor}         
\usepackage{amsmath} 
\usepackage{graphicx}
\usepackage{multirow}
\usepackage{wrapfig}
\usepackage{booktabs}
\usepackage{tabularx}
\usepackage{float}
\usepackage{enumitem}
\hypersetup{
    colorlinks=true,
    citecolor=black,
    linkcolor=black,
    urlcolor=black
}
\newtheorem{proposition}{Proposition}

\newtheorem{assumption}{Assumption}

\title{Flow-Guided Target-Space Alignment via Path Consistency}

\author{
\textbf{Ruizhi Yuan}$^{1}$\thanks{Equal contribution.} \quad
\textbf{Zeqiu Yu}$^{2}$\footnotemark[1] \quad
\textbf{Wei Gao}$^{3}$ \quad
\textbf{Wei Chen}$^{1,4}$ \quad
\textbf{Xiwei Tang}$^{5}$\thanks{Equal supervision.} \quad
\textbf{Lu Tang}$^{1}$\footnotemark[2] \\[0.5em]
$^{1}$Department of Biostatistics and Health Data Science,
University of Pittsburgh \\
$^{2}$Department of Electrical and Computer Engineering,
University of Virginia \\
$^{3}$Department of Electrical and Computer Engineering,
University of Pittsburgh \\
$^{4}$Department of Pediatrics,
University of Pittsburgh \\
$^{5}$Department of Mathematical Sciences,
University of Texas at Dallas \\[0.3em]
{\small\texttt{xiwei.tang@utdallas.edu, lutang@pitt.edu}}
}

\begin{document}

\maketitle

\begin{abstract}
Many multimodal alignment problems require mapping a source modality into a fixed space defined by a target modality, where the target space is not merely an intermediate embedding but is used directly for downstream analysis. In this setting, successful alignment should preserve both the global structure of the target distribution and the instance-level correspondence observed in paired source--target training data. Existing objectives typically address only part of this goal: contrastive losses encourage relative similarity, distribution-matching losses improve marginal agreement, and pointwise regression fits each source sample to its paired target without explicitly modeling target-space path geometry. We propose FlowAlign, a flow-guided target-space alignment method that repurposes conditional flow matching as a training-time alignment signal rather than an inference-time transport model. FlowAlign uses the learned flow field to guide the alignment of source-side representations with the fixed target space, encouraging them to preserve paired correspondence while remaining compatible with the global target-space structure. We further provide theoretical analysis showing how FlowAlign helps improve paired correspondence by controlling endpoint accuracy and capturing local target-space geometry.
Experiments on synthetic data, paired single-cell multiome benchmarks, and image--text alignment demonstrate that FlowAlign improves instance level correspondence while preserving target-space structure across different modality settings. Code and instructions for reproducing our experiments are available at \url{https://github.com/yrzzz/FlowAlign}.
\end{abstract}

\section{Introduction}

Many multimodal alignment problems are solved by learning a shared representation space.
This strategy is natural when neither modality has a privileged analysis space.
However, recent work also shows the value of keeping one representation system fixed and adapting another modality to it.
In vision-language learning, locked or frozen-model approaches train one modality to become compatible with a strong pretrained tower or language-model interface~\citep{zhai2022lit,tsimpoukelli2021multimodal,mokady2021clipcap,li2023blip,manas2023mapl}.
Similar source-to-target adaptation appears in brain decoding, where fMRI activity is mapped into frozen CLIP or diffusion latent spaces for retrieval and reconstruction~\citep{scotti2023reconstructing,ma2025brainclip,ozcelik2023natural}, and in generative personalization, where user-provided visual concepts are represented as new tokens or textual embeddings in frozen text-to-image models~\cite{gal2022image,wei2023elite}.

In some scientific domains, the target representation space is not merely an interface for prediction or generation, but the output analysis space itself.
Mapped source samples should therefore preserve the geometry that makes the target space useful, including neighborhoods, clusters, annotations, and, when paired target observations are meaningful, sample-level correspondence.
Paired single-cell multiome analysis provides a representative example.
Here, scATAC-seq and scRNA-seq are two modalities measured from the same cells, and a common goal is to map ATAC profiles into an RNA-defined representation space for downstream analysis~\citep{stuart2019comprehensive,hao2024dictionary}.
Because the two profiles are paired at the cell level, this task is not only about matching the overall RNA distribution or recovering broad cell-type neighborhoods; it also requires preserving RNA-space geometry and paired-cell correspondence~\citep{cao2022multi,hu2024scpair,xiao2024benchmarking}. This exposes a broader limitation of existing alignment objectives: contrastive, distribution-matching, and pointwise regression losses emphasize relative similarity, marginal coverage, and paired fitting, respectively, but do not jointly preserve target-space structure and sample-level correspondence.

We revisit flow matching \cite{lipman2023flow,tong2023improving} from this perspective.
Flow matching is typically used to learn a velocity field for transport or generation.
Here, we use the learned field differently: not as an inference-time generator, but as a training-time guide for aligning source-side representations with the fixed target space.
The intuition is that conditional flow matching defines paths between paired samples and learns velocity fields along them.
In target-space alignment, an aligned representation may lie in the target space but still fail to match the correct paired target.
Such a mismatch connects the source representation to an incorrect target-space location, so the resulting path requires velocities that are inconsistent with those learned from true paired examples.
This motivates a path-consistency criterion that uses velocity agreement as a training signal for alignment.

We therefore introduce FlowAlign, a flow-guided target-space alignment method based on this criterion.
FlowAlign learns a source-conditioned velocity field from paired source and target samples and uses the resulting path-consistency loss to guide the alignment of source-side representations toward the fixed target space.

We analyze this path-consistency criterion from both pairwise and geometric perspectives.
At the pair level, the analysis connects the criterion to endpoint accuracy; at the geometric level, it shows how accurate target-space alignment can support neighborhood stability.
Together, these results suggest that path consistency can support both paired correspondence and target-space structural stability.
We then validate these implications across three experimental settings.
In a controlled synthetic task, FlowAlign maintains correct source-to-target matching while recovering target distributional structure.
On paired single-cell multiome datasets, FlowAlign maps ATAC-derived representations into fixed RNA target spaces, improving paired-cell matching and preserving RNA-defined biological structure.
We further evaluate FlowAlign on a paired image--text alignment task, demonstrating that its benefits extend beyond single-cell modalities.

In summary, FlowAlign makes three contributions:
\begin{itemize}[leftmargin=*, itemsep=0.25em, topsep=0.25em]
\item We study a latent space alignment setting in which one modality is aligned into the fixed representation space of a target modality, requiring both global target-space structure and paired instance-level correspondence to be preserved.
\item We propose FlowAlign, a flow-guided path-consistency criterion that repurposes conditional flow matching as a training-time alignment guide and can be used either as a standalone objective or as a plug-in regularizer for existing task-specific models.
\item We provide theoretical analysis of endpoint control and local target-space geometry, and validate FlowAlign on synthetic, paired single-cell multiome, and image--text alignment tasks, demonstrating improved correspondence while preserving target-space structure.
\end{itemize}

\section{Related Work}

\paragraph{Multimodal Alignment objectives and domain adaptation.}
Existing alignment objectives capture different notions of agreement between modalities or domains. 
Contrastive objectives use paired data to enforce relative similarity between matched and unmatched samples~\citep{oord2018representation,wang2020understanding}. 
Distribution-level objectives instead align source and target marginals, for example through adversarial domain confusion~\citep{ganin2016domain}, maximum mean discrepancy~\citep{long2015learning}, correlation alignment~\citep{sun2016deep}, or optimal-transport couplings~\citep{courty2017joint}. 
Pointwise regression directly fits paired target endpoints when paired supervision is available. 
Together, these objectives emphasize complementary notions of alignment.

\paragraph{Single-cell multi-omics integration.}
Single-cell multi-omics methods integrate modalities such as scRNA-seq and scATAC-seq through several modeling strategies. 
Reference- and bridge-mapping methods transfer information across modalities using a multiome reference, with Seurat v5 bridge integration as a representative example~\citep{hao2024dictionary}. 
Graph-based approaches such as GLUE incorporate prior regulatory relationships through guidance graphs that connect features across omics layers~\citep{cao2022multi}. 
Deep generative and shared-latent methods, including MultiVI and Cobolt, model multiple modalities through probabilistic or joint latent representations~\citep{ashuach2023multivi,gong2021cobolt}. 
Other approaches emphasize transfer learning, trajectory preservation, paired-cell correspondence, or optimal transport, including scJoint, scDART, scPair, bindSC, and uniPort~\citep{lin2022scjoint,zhang2022scdart,hu2024scpair,dou2022bi,cao2022unified}.

\paragraph{Flow-based transport and generative modeling.}
Neural ODEs and continuous normalizing flows model data through continuous-time dynamics, often using ODE integration for inference or likelihood computation~\citep{chen2018neuralode,grathwohl2019ffjord}. 
Flow matching and related continuous-time generative models learn velocity fields that transport samples between distributions, often through ODE- or SDE-based inference~\citep{lipman2023flow,tong2023improving,liu2023flow}. 
Related frameworks include stochastic interpolants~\citep{albergo2023stochastic}, simulation-free Schr\"odinger bridges~\citep{tong2023simulation}, action matching~\citep{neklyudov2023action}, and flow-map matching~\citep{boffi2025flow}. 
These methods primarily use learned dynamics for generation, transport, likelihood evaluation, or simulation.
Related approaches include GENOT~\citep{klein2024genot}, which uses flow matching for unpaired neural optimal transport, and score distillation sampling~\citep{poole2022dreamfusion}, which uses pretrained diffusion scores to guide optimization.

\section{Preliminaries}

\subsection{Source-to-Target Alignment in a Fixed Target Space}
\label{sec:prelim_fixed_target}

We consider paired source and target observations 
\(\mathcal D=\{(a_i,b_i)\}_{i=1}^n\), where \(a_i\in\mathcal A\) and \(b_i\in\mathcal B\) denote samples from the source and target domains, respectively. 
Fixed representation pipelines map the two domains to vectors \(s_i=f_S(a_i)\in\mathbb R^d\) and \(t_i=f_T(b_i)\in\mathbb R^d\). 
The source representation \(s_i\) may come from PCA, a pretrained source encoder, or a pretrained feature space followed by a fixed projection to dimension \(d\) determined by the target space. 
The equality of dimensions is used only to define paths in a common computational ambient space. 
We do not assume that \(s_i\) is already expressed in the target coordinate system or aligned with the target representations. 
The fixed target coordinate system is defined by the target representations \(t_i\), which form the target embedding space \(\mathcal T\subseteq\mathbb R^d\) and are kept fixed throughout training and evaluation.

The goal is to produce a target-space representation for each source sample. 
At inference time, only the source side is available, and the trainable source-side model outputs \(\hat t_\theta\in\mathbb R^d\). 
The output \(\hat t_\theta\) is interpreted in the same coordinate system as the target representations, so that distances, neighborhoods, clusters, and annotations are evaluated within \(\mathcal T\).

\subsection{Conditional Flow Matching}
\label{sec:prelim_cfm}

Flow matching \cite{lipman2023flow} parameterizes a time-dependent velocity field 
\(v_\psi:\mathbb R^d\times[0,1]\to\mathbb R^d\). 
A trajectory \(x_\tau\) generated by this field follows the ODE \(d x_\tau/d\tau=v_\psi(x_\tau,\tau)\), where \(\tau\in[0,1]\). 
Conditional flow matching (CFM) avoids directly constructing the marginal velocity field by regressing against conditional velocities along tractable probability paths \cite{tong2023improving}. 
Let \(c\) denote a condition, \(p_\tau(\cdot\mid c)\) a conditional path, and \(u_\tau(\cdot\mid c)\) its conditional velocity. 
The CFM objective is
\[
\mathcal L_{\mathrm{CFM}}(\psi)
=
\mathbb E_{\tau\sim\mathcal U[0,1],\,c,\,x_\tau\sim p_\tau(\cdot\mid c)}
\left[
\left\|
v_\psi(x_\tau,\tau)-u_\tau(x_\tau\mid c)
\right\|_2^2
\right].
\]

A common instantiation uses endpoint pairs to define linear conditional paths. Given two endpoints \(x_0,x_1\in\mathbb R^d\), the linear path is \(x_\tau=(1-\tau)x_0+\tau x_1\), with conditional velocity \(u_\tau(x_\tau\mid x_0,x_1)=x_1-x_0\). 
Substituting this path into the CFM objective yields the standard linear-path flow matching objective, corresponding to rectified flow under the chosen endpoint coupling\cite{liu2023flow}:
\begin{equation}
\label{eq:linear_fm}
\mathcal L_{\mathrm{RF}}(\psi)
=
\mathbb E_{\tau\sim\mathcal U[0,1],\,(x_0,x_1)}
\left[
\left\|
v_\psi((1-\tau)x_0+\tau x_1,\tau)
-
(x_1-x_0)
\right\|_2^2
\right].
\end{equation}

This loss form will be used later to define source-conditioned paths in the fixed target space.




\section{FlowAlign: Flow-Guided Target-Space Alignment}
\label{sec:method}

We propose FlowAlign, a flow-guided target-space alignment method that uses conditional flow matching to define a path-consistency score for training. 
The method has two stages. 
First, we learn a source-conditioned velocity field along paired source--target interpolation paths. 
Second, we freeze this field and use it to define a path-consistency score for candidate representations of source samples in the target space.
At inference time, the velocity field and path sampling are not used; the trained source-side model directly outputs representations in the fixed target space.

\subsection{Learning a Source-Conditioned Velocity Field}
\label{sec:method_velocity_field}

In Stage 1, we first learn a source-conditioned velocity field by instantiating the standard linear-path flow matching objective in Eq.~\eqref{eq:linear_fm} on paired source and target representations. For a paired example \((s,t)\), the fixed source representation \(s\in\mathbb R^d\) defines the source-conditioned start center, and \(t\in\mathbb R^d\) is the paired target endpoint. 
Using \(s\) as the start center does not assume that it is already a valid target-space representation; it only specifies where source-conditioned paths begin. 
We define \(p_0(\cdot\mid s)=\mathcal N(s,\sigma^2 I)\), where \(\sigma\) controls the local neighborhood of start points around \(s\).

Given \((s,t)\), we sample \(x_0\sim p_0(\cdot\mid s)\) and \(\tau\sim\mathcal U[0,1]\). 
We form the linear path \(x_\tau=(1-\tau)x_0+\tau t\), whose target velocity is \(t-x_0\). 
The source-conditioned velocity field \(v_\psi\): \(\mathbb R^d\times[0,1]\times\mathbb R^d\to\mathbb R^d\) is trained by
\begin{equation}
\label{eq:field_loss}
\mathcal L_{\mathrm{field}}(\psi)
=
\mathbb E_{(s,t),\,x_0\sim p_0(\cdot\mid s),\,\tau\sim\mathcal U[0,1]}
\left[
\left\|
v_\psi((1-\tau)x_0+\tau t,\tau;s)
-
(t-x_0)
\right\|_2^2
\right].
\end{equation}

The conditioning on \(s\) makes the field source-specific, while the local start distribution trains the field on a neighborhood of paths rather than a single deterministic segment, analogous to noisy conditional paths in flow matching and stochastic interpolants~\citep{tong2023improving,albergo2023stochastic}.
The special case \(\sigma=0\) reduces to the deterministic path from \(s\) to \(t\).
After optimizing Eq.~\eqref{eq:field_loss}, we freeze \(v_\psi\). 

\subsection{Path-Consistency Score}
\label{sec:method_path_critic}

\begin{figure}[t]
  \centering
  \includegraphics[width=1\textwidth]{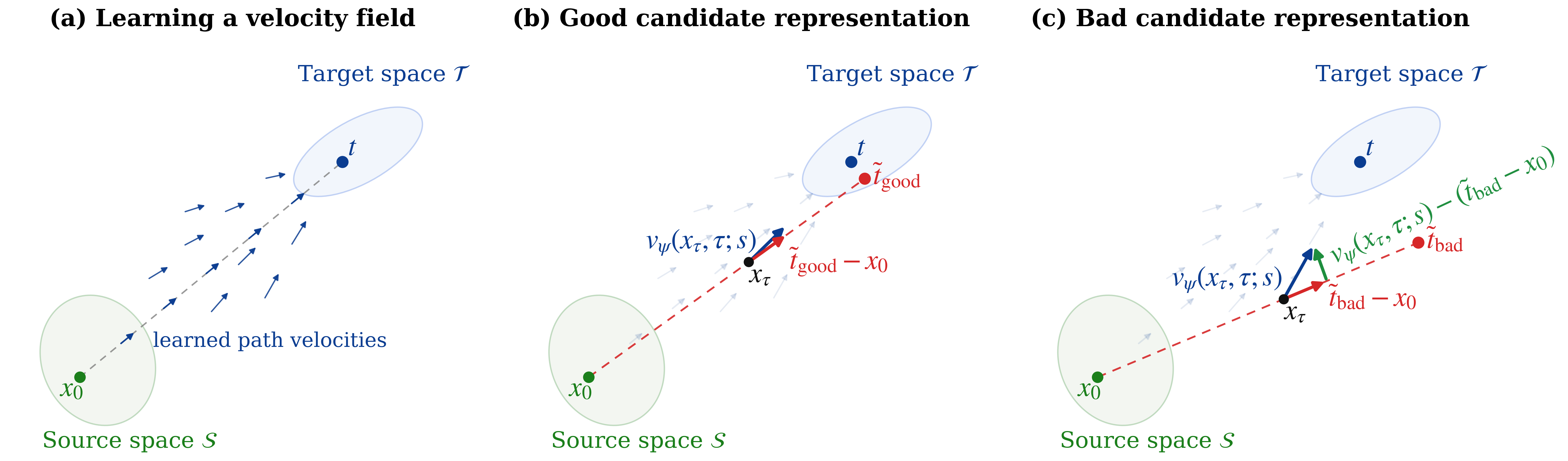}
  \caption{
  Path-consistency score.
  (a) Field learning on paired source-to-target interpolation paths.
  (b) A compatible candidate representation yields a small velocity residual.
  (c) An incompatible candidate representation yields a large velocity residual.
  }
  \label{fig:path_consistency_critic}
\end{figure}

We use the frozen source-conditioned velocity field \(v_\psi\) to score candidate target-space representations of source samples in Stage 2. 
Figure~\ref{fig:path_consistency_critic} illustrates the intuition: compatible candidate representations induce paths whose velocities agree with the frozen field, whereas incompatible representations produce large residuals. 
Given a fixed source representation \(s\) and a candidate representation in the target space \(\tilde t\in\mathbb R^d\), we sample \(x_0\sim p_0(\cdot\mid s)\) and form the candidate path \(x_\tau^{\tilde t}=(1-\tau)x_0+\tau\tilde t\). 
We define the path-consistency score as
\begin{equation}
\label{eq:path_score}
\mathcal C_\psi(\tilde t;s)
=
\mathbb E_{x_0\sim p_0(\cdot\mid s),\,\tau\sim\mathcal U[0,1]}
\left[
\left\|
v_\psi((1-\tau)x_0+\tau\tilde t,\tau;s)
-
(\tilde t-x_0)
\right\|_2^2
\right].
\end{equation}
A small value of \(\mathcal C_\psi(\tilde t;s)\) in Eq.~\eqref{eq:path_score} indicates that the candidate representation in the target space induces a path whose local velocity is consistent with the learned source-conditioned dynamics.

The same residual plays different roles in the two stages. 
During field training, the endpoint is the observed paired target \(t\), and the residual is minimized with respect to \(\psi\). 
Here, \(v_\psi\) is frozen and the candidate representation \(\tilde t\) is variable, so the score evaluates candidate representations rather than using the field as an inference-time ODE generator.

\subsection{Training with the Path-Consistency Loss}
\label{sec:method_training}

For Stage 2, we now use the frozen path-consistency loss to train the source-side model. 
For each source sample, let \(\hat t_\theta\in\mathbb R^d\) denote the candidate target-space representation produced by the trainable model. 
The loss is evaluated with the same fixed source representation \(s\) used during Stage-1 field training. 
Thus, \(s\) serves as the condition for the frozen field, while \(\hat t_\theta\) is the output optimized to lie in the fixed target coordinate system.

When using path consistency as the standalone training signal, we optimize  Eq.~\eqref{eq:path_loss}
\begin{equation}
\label{eq:path_loss}
\mathcal L_{\mathrm{path}}(\theta)
=
\mathbb E_s
\left[
\mathcal C_\psi(\hat t_\theta;s)
\right],
\end{equation}
where \(\mathcal C_\psi\) is evaluated by sampling \(x_0\sim p_0(\cdot\mid s)\) and \(\tau\sim\mathcal U[0,1]\). 
In the standalone-mapper setting, \(\hat t_\theta=g_\theta(s)\). 
If the trainable model includes a projection head, encoder, or autoencoder latent output, \(\hat t_\theta\) denotes that optimized target-space output, while the frozen velocity field remains conditioned on the fixed source representation \(s\). 
This avoids changing the conditioning input of \(v_\psi\) after the field has been trained. 
Although \(t\) does not appear explicitly in the Stage-2 loss, the score remains paired-data informed because \(v_\psi\) was trained in Stage 1 on paired paths ending at the paired target representation.

The same score can also be combined with a base objective,
\[
\min_\theta
\quad
\mathcal L_{\mathrm{base}}(\theta)
+
\lambda
\mathcal L_{\mathrm{path}}(\theta),
\]
where \(\mathcal L_{\mathrm{base}}\) may be an endpoint loss, reconstruction loss, contrastive loss, label-prediction loss, or another task-specific objective. 
In this form, the path-consistency loss acts as a plug-in target-space alignment regularizer.

During this stage, \(v_\psi\) is frozen in parameters but not detached from its input. 
The candidate path point \(x_\tau^{\hat t_\theta}=(1-\tau)x_0+\tau\hat t_\theta\) depends on \(\theta\), so the gradients of \(\mathcal C_\psi(\hat t_\theta;s)\) propagate through the candidate path and update the source-side model.

At inference time, the velocity field and path sampling are discarded. 
A source sample is mapped to the fixed target space by a single forward pass through the trained source-side model.

\subsection{Properties of the Path-Consistency Score}
\label{sec:method_theory}

The path-consistency score is built on a simple link between endpoint error and path error. 
For a true paired target \(t\) and a candidate aligned representation of the source \(\tilde t\), define the paired path 
\(x_\tau^t=(1-\tau)x_0+\tau t\) and the candidate path 
\(x_\tau^{\tilde t}=(1-\tau)x_0+\tau \tilde t\). 
With \(\delta=\tilde t-t\), these paths satisfy
\begin{equation}
\label{eq:path_shift}
x_\tau^{\tilde t}-x_\tau^t=\tau\delta .
\end{equation}
Eq.~\eqref{eq:path_shift} shows that changing the aligned representation changes every intermediate point on the candidate path. 
The path consistency exploits this fact by checking whether the velocity required by the candidate path agrees with the frozen source-conditioned field learned from paired paths.

We formalize three consequences of this observation. 
First, a low path-consistency value controls endpoint error up to the paired-path residual of the frozen velocity field.
Second, locally, the path consistency behaves as a learned source-conditioned metric over endpoint perturbations. 
Third, source conditioning reduces velocity ambiguity when different source instances require different directions through similar target-space locations.

\paragraph{Controlling endpoint error.}
Fix a paired example \((s,t)\) and a candidate endpoint \(\tilde t\). 
Let \(x_0\sim p_0(\cdot\mid s)\) and \(\tau\sim\mathcal U[0,1]\). 
We define the paired-path residual of the frozen field as
\begin{equation}
\label{eq:paired_residual}
\varepsilon_\psi(s,t)^2
=
\mathbb E_{x_0,\tau}
\left[
\left\|
v_\psi((1-\tau)x_0+\tau t,\tau;s)
-
(t-x_0)
\right\|_2^2
\right].
\end{equation}
The residual in Eq.~\eqref{eq:paired_residual} measures how accurately the Stage-1 field matches the linear-path velocity on the paired source--target path.

\begin{assumption}[Local smoothness of the frozen field]
\label{assump:lipschitz_field}
For a fixed source \(s\), the frozen field is locally smooth along the paired and candidate interpolation paths. 
Specifically, there exists \(L\ge 0\) such that, for all path points \(x,x'\) and all \(\tau\in[0,1]\),
\[
\|v_\psi(x,\tau;s)-v_\psi(x',\tau;s)\|_2
\le
L\|x-x'\|_2 .
\]
\end{assumption}

\begin{proposition}[Approximate endpoint control]
\label{prop:endpoint_control}
Fix a paired example \((s,t)\) and a candidate representation \(\tilde t\). 
Under Assumption~\ref{assump:lipschitz_field},
\begin{equation}
\label{eq:endpoint_control}
\|\tilde t-t\|_2
\le
\frac{
\sqrt{\mathcal C_\psi(\tilde t;s)}
+
\varepsilon_\psi(s,t)
}{
\sqrt{\kappa(L)}
},
\qquad
\kappa(L)=\int_0^1 (1-L\tau)_+^2\,d\tau .
\end{equation}
\end{proposition}


Proposition~\ref{prop:endpoint_control} shows that a candidate representation with a small path-consistency score is close to the paired target whenever the frozen field is accurate along the paired interpolation path. 
Equivalently, the reliability of the score depends on two quantities: the candidate score \(\mathcal C_\psi(\tilde t;s)\) and the paired-path residual \(\varepsilon_\psi(s,t)\) in Eq.~\eqref{eq:paired_residual}. 
In the realizable case where \(\varepsilon_\psi(s,t)=0\), Eq.~\eqref{eq:endpoint_control} reduces to
\[
\|\tilde t-t\|_2
\le
\kappa(L)^{-1/2}\sqrt{\mathcal C_\psi(\tilde t;s)} .
\]
In particular, \(\mathcal C_\psi(\tilde t;s)=0\) implies \(\tilde t=t\). 
The proof follows by comparing the candidate and paired paths using Eq.~\eqref{eq:path_shift} and applying the reverse triangle inequality together with the Lipschitz condition. 

For a trained source-side model with target-space output \(\hat t_\theta\), Proposition~1 yields the average bound
\[
\mathbb E_s\|\hat t_\theta-t\|_2^2
\le
\frac{2}{\kappa(L)}
\left(
\mathbb E_s[\mathcal C_\psi(\hat t_\theta;s)]
+
\mathbb E_s[\varepsilon_\psi(s,t)^2]
\right),
\]
assuming the same Lipschitz constant \(L\) holds over the evaluated examples. 
This bound makes explicit the two-stage structure of the method: the final mapping error is controlled by both the path-consistency loss optimized in Stage 2 and the field quality obtained in Stage 1.

\paragraph{Local metric interpretation.}
The path-consistency score also has a local geometric interpretation. 
Let \(\delta=\tilde t-t\), and assume that \(v_\psi\) is differentiable in its state argument near the paired path. 
Since \(x_\tau^{\tilde t}-x_\tau^t=\tau\delta\), a first-order expansion around \(x_\tau^t\) gives
\[
v_\psi(x_\tau^{\tilde t},\tau;s)
-
v_\psi(x_\tau^t,\tau;s)
\approx
\tau J_\psi(x_\tau^t,\tau;s)\delta,
\]
where \(J_\psi=\nabla_x v_\psi\). 
In the regime where the paired-path residual is small, the path-consistency score satisfies the local approximation
\begin{equation}
\label{eq:local_metric}
\mathcal C_\psi(\tilde t;s)
\approx
\delta^\top A_\psi(s)\delta,
\qquad
A_\psi(s)
=
\mathbb E_{x_0,\tau}
\left[
(I-\tau J_\psi)^\top
(I-\tau J_\psi)
\right],
\end{equation}
with the Jacobian evaluated along the paired path. 
Eq.~\eqref{eq:local_metric} shows that, near the paired target, the path-consistency score behaves like a learned source-conditioned quadratic penalty on endpoint errors. 
Unlike endpoint MSE, which uses the fixed isotropic form \(\delta^\top I\delta\), the path-consistency score uses the field-induced matrix \(A_\psi(s)\), allowing different endpoint-error directions to be penalized according to the local source-conditioned geometry of the frozen velocity field.

This interpretation also clarifies the degenerate case. 
If \(v_\psi(x,\tau;s)\) has little dependence on \(x\) or \(\tau\), then \(J_\psi\approx 0\) and \(A_\psi(s)\approx I\), so the score becomes close to an endpoint-style penalty. 
The benefit of the path-consistency score therefore comes from learning a nontrivial source-conditioned velocity field over the target coordinate space.

\paragraph{Role of source conditioning.}
Finally, source conditioning reduces ambiguity in the velocity regression problem. 
Let \(X_\tau\) denote a path point and let \(V\) denote its target velocity. 
For squared velocity regression, the optimal unconditioned field predicts \(\mathbb E[V\mid X_\tau,\tau]\), with irreducible error \(\mathrm{Tr}(\mathrm{Var}(V\mid X_\tau,\tau))\) where \(\mathrm{Tr}\) denotes the matrix trace. 
A source-conditioned field instead predicts \(\mathbb E[V\mid X_\tau,\tau,S]\), with irreducible error \(\mathrm{Tr}(\mathrm{Var}(V\mid X_\tau,\tau,S))\). 
Since conditioning reduces variance,
\begin{equation}
\label{eq:conditioning_variance}
\mathbb E\!\left[
\mathrm{Tr}(\mathrm{Var}(V\mid X_\tau,\tau,S))
\right]
\le
\mathbb E\!\left[
\mathrm{Tr}(\mathrm{Var}(V\mid X_\tau,\tau))
\right].
\end{equation}
Eq.~\eqref{eq:conditioning_variance} formalizes why conditioning on \(s\) can reduce velocity ambiguity. 
The inequality is strict when the source representation disambiguates different velocities passing through similar path locations. 
This explains why the path-consistency score is conditioned on \(s\): without source conditioning, intersecting or nearby paths with different target velocities are averaged, whereas a source-conditioned field can assign different directions to the same target-space location depending on the source instance.

Detailed proofs and neighborhood-stability implications are provided in Appendix~\ref{app:theory}.

\section{Experiments}

We evaluate FlowAlign on synthetic data, paired single-cell multiome benchmarks, and image--text alignment.
The synthetic experiment compares FlowAlign with endpoint regression, contrastive alignment, and distribution matching.
The PBMC and BMMC benchmarks evaluate source-only mapping of ATAC-derived representations into fixed RNA target spaces, comparing FlowAlign with both established multi-omics integration methods and architecture-matched paired-supervision baselines.
We further evaluate image-to-text mapping on CUBICC to assess whether the method generalizes beyond single-cell modalities.
Finally, ablations examine local-start sampling, source conditioning, and the use of the learned field as a training-time score rather than an inference-time transport map.
Implementation details are provided in Appendix~\ref{app:experiment_details}.

\subsection{Synthetic target-space alignment}
\label{sec:exp_synthetic}

\begin{figure}
  \centering
  \includegraphics[width=0.8\textwidth]{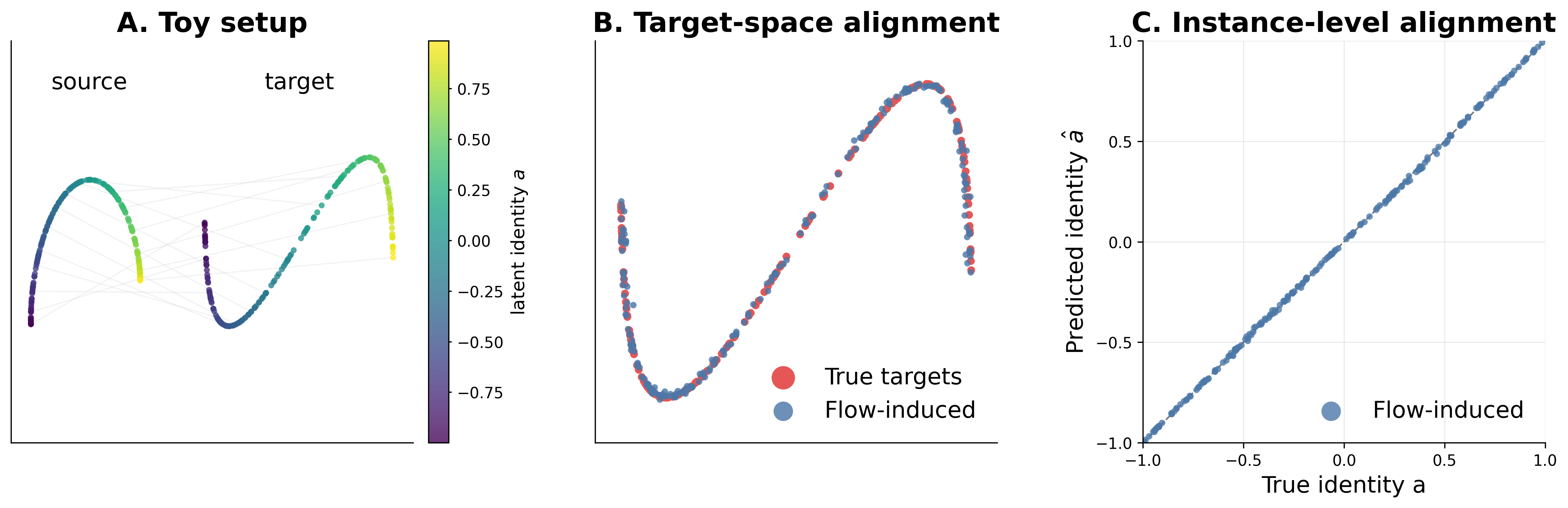}
  \caption{Synthetic fixed-target alignment.
(A) Source and target manifolds colored by latent coordinate \(a\).
(B) Predicted target-space points.
(C) Predicted versus true latent positions.
}
  \label{fig:synthetic}
\end{figure}

For synthetic data, each sample is generated from a latent identity \(a\in[-1,1]\). 
The target point \(t(a)\in\mathbb R^2\) lies on a fixed two-dimensional trajectory, and the source point \(s(a)\in\mathbb R^2\) is a shifted and distorted version of the same trajectory:
\[
\begin{aligned}
t(a)
&=
\begin{bmatrix}
0.95a+0.28\sin(\pi a)\\
0.78\sin(1.15\pi a)+0.16a
\end{bmatrix},
\qquad
s(a)
=
t(a)
+
\begin{bmatrix}
-0.75+0.06\cos(2\pi a)\\
-0.28-0.05\sin(\pi a)
\end{bmatrix}
+\epsilon ,
\end{aligned}
\]
where \(\epsilon\sim\mathcal N(0,\sigma_s^2I)\). 
The task is to map each source point \(s(a)\) to its paired target endpoint \(t(a)\) in the fixed target coordinate system. 
Paired source--target samples are generated with \(a\) sampled uniformly over \([-1,1]\). 
For evaluating distributional agreement with MMD, we additionally sample target points from a shifted marginal distribution over \(a\), with higher density near the middle of the trajectory. 
Since \(s(a)\) is already two-dimensional, we use it as the source-conditioned start center and set \(p_0(\cdot\mid s)=\mathcal N(s,\sigma^2I)\). Figure~\ref{fig:synthetic}A shows the source data and target data in the 2D space. 

We compare four methods: direct endpoint regression with an MSE loss, a distribution-wise baseline trained with a Maximum Mean Discrepancy (MMD) loss, a fixed-target InfoNCE \cite{oord2018representation} baseline, and FlowAlign trained with the path-consistency objective in Eq.~\eqref{eq:path_loss}. 
We evaluate them using three complementary metrics. 
MMD measures how well the predicted points match the target distribution. 
Endpoint MSE measures paired coordinate accuracy against the true target point \(t(a_i)\). 
To further evaluate whether each source sample is mapped to the correct location along the target trajectory, we compute a latent-position error by projecting each prediction back to the trajectory:
$\hat a_i=\arg\min_{a'\in[-1,1]}\|\hat t_i-t(a')\|_2,
\mathrm{MAE}_a
=
\frac{1}{n}\sum_i|\hat a_i-a_i|.
$

\begin{wraptable}{r}{0.57\textwidth}
\vspace{-1.8em}
\caption{
Synthetic experiment results on held-out paired samples. Lower values indicate better performance.
}
\centering
\label{tab:synthetic}
\small
\begin{tabularx}{0.57\textwidth}{lccc}
\toprule
\textbf{Method}
& \textbf{MMD} $\downarrow$
& \textbf{MSE} $\downarrow$
& \textbf{MAE\(_a\)} $\downarrow$ \\
\midrule
MSE reg.
& $8.13{\times}10^{-2}$
& $3.54{\times}10^{-4}$
& $5.61{\times}10^{-3}$ \\
MMD align.
& $\mathbf{6.99{\times}10^{-3}}$
& $1.50{\times}10^{-1}$
& $1.68{\times}10^{-1}$ \\
InfoNCE
& $2.30{\times}10^{-1}$
& $3.79{\times}10^{-1}$
& $9.78{\times}10^{-2}$ \\
FlowAlign
& $7.79{\times}10^{-2}$
& $\mathbf{2.87{\times}10^{-4}}$
& $\mathbf{5.10{\times}10^{-3}}$ \\
\bottomrule
\end{tabularx}
\end{wraptable}

Figure~\ref{fig:synthetic} and Table~\ref{tab:synthetic} summarize the results. 
The baselines show complementary failure modes: MMD alignment best matches the target marginal but has large paired errors, while InfoNCE performs poorly at predicting the correct location in the fixed target space. 
FlowAlign achieves the lowest endpoint MSE and \(\mathrm{MAE}_a\), indicating improved paired alignment accuracy while maintaining reasonable target-space coverage. Baseline visualizations are provided in Appendix~\ref{app:toy_exp}.

\subsection{Single-cell multiome target-space alignment}
\label{sec:exp_single_cell}

We evaluate FlowAlign on public single-cell multiome data, where chromatin accessibility measured by ATAC-seq and gene expression measured by RNA-seq are profiled from the same cells.
We use PBMC \cite{10x_pbmc_multiome} and BMMC \cite{geo_gse194122,lance2022multimodal} as representative peripheral blood and bone marrow benchmarks. 
Both modalities are processed using standard pipelines: ATAC profiles are embedded with TF--IDF/LSI and RNA profiles with PCA, yielding 25-dimensional source and fixed target representations~\cite{stuart2021signac}. In our setting, ATAC latent is the source modality and RNA latent defines the target analysis space. 

For each dataset, cells are split at the paired-cell level into training, validation, and held-out test sets. 
Models are trained on paired ATAC--RNA cells and evaluated under source-only inference, where only ATAC-derived test representations are provided and paired RNA embeddings are used only for evaluation. 
In this experiment, FlowAlign uses the standalone path-consistency objective in Eq.~\eqref{eq:path_loss} to train a mapper without any additional terms.
We compare FlowAlign with representative single-cell integration methods, including Seurat v5 Bridge Integration~\citep{hao2024dictionary}, GLUE~\citep{cao2022multi}, MultiVI~\citep{ashuach2023multivi}, and Cobolt~\citep{gong2021cobolt}.
To provide controlled comparisons under the same paired supervision, we additionally evaluate direct MSE regression, MSE+MMD, MSE+InfoNCE, and MSE+neighborhood preservation using the same mapper architecture, data splits, and training protocol.
The neighborhood-preservation baseline encourages the mapped representations to preserve pairwise distances along local neighbor edges defined in the target space.
For the established integration methods, metrics are computed in their native aligned spaces, whereas the paired regression baselines and FlowAlign are evaluated directly in the fixed RNA PCA target space.
The former provide application-level comparisons, while the latter isolate the effect of the alignment objective under matched supervision.

We evaluate paired-cell correspondence using FOSCTTM~\citep{liu2019jointly} and exact cell match (R@1), and biological structure using label transfer accuracy (LTA), NMI, ARI, and ASW-O~\citep{xiao2024benchmarking}. 
Lower FOSCTTM is better; higher values are better for all other metrics. Detailed metric definitions are provided in Appendix~\ref{app:evaluation_metrics}.
Table~\ref{tab:main_target_space} summarizes the results.
Across both datasets, FlowAlign achieves the lowest FOSCTTM and highest exact cell-match rate among the evaluated methods.
Compared with the architecture-matched paired baselines, FlowAlign also obtains higher NMI and ARI, indicating improved preservation of RNA-defined biological structure.
LTA is highest for FlowAlign on both datasets, although the improvement over the strongest paired baseline on PBMC is small.
ASW-O remains competitive, but is not uniformly the highest across methods.
These results suggest that path consistency improves fine-grained paired correspondence while maintaining target-space biological organization.
Mean \(\pm\) standard deviation over three independent runs is reported in Appendix~\ref{app:seed_variance}, with additional retrieval and biological-structure metrics in Appendix~\ref{app:sc_supp}.

Figure~\ref{fig:test_umap} visualizes the held-out BMMC predictions using UMAP~\cite{mcinnes2018umap}.
The mapped ATAC embeddings overlap the RNA reference geometry and reproduce major RNA-defined cell-type neighborhoods, consistent with the quantitative improvements in Table~\ref{tab:main_target_space}. A UMAP visualization of the held-out PBMC predictions is provided in Figure~\ref{fig:pbmc_umap} in Appendix~\ref{app:sc_supp}.
\begin{table}[t]
\centering
\caption{
Performance comparison on held-out single-cell multiome alignment.
All methods are evaluated using the same paired-cell and biological-structure metrics.
}
\renewcommand{\arraystretch}{1.25}
\label{tab:main_target_space}
\resizebox{\textwidth}{!}{
\begin{tabular}{lcccccccccccc}
\toprule
\multirow{2}{*}{\textbf{Method}}
& \multicolumn{6}{c}{\textbf{PBMC}}
& \multicolumn{6}{c}{\textbf{BMMC}} \\
\cmidrule(lr){2-7}
\cmidrule(lr){8-13}

& \textbf{FOSCTTM} $\downarrow$
& \textbf{Cell match} $\uparrow$
& \textbf{LTA} $\uparrow$
& \textbf{NMI} $\uparrow$
& \textbf{ARI} $\uparrow$
& \textbf{ASW-O} $\uparrow$
& \textbf{FOSCTTM} $\downarrow$
& \textbf{Cell match} $\uparrow$
& \textbf{LTA} $\uparrow$
& \textbf{NMI} $\uparrow$
& \textbf{ARI} $\uparrow$
& \textbf{ASW-O} $\uparrow$ \\
\midrule
\multicolumn{13}{l}{\textit{Integration methods (native aligned spaces)}} \\
Seurat v5
& 0.058 & 0.019 & 0.850 & 0.783 & 0.785 & \textbf{0.995}
& 0.045 & 0.008 & 0.756 & 0.714 & 0.655 & 0.978 \\

GLUE
& 0.044 & 0.016 & 0.914 & 0.802 & 0.813 & 0.979
& 0.056 & 0.015 & 0.745 & 0.701 & 0.689 & 0.981 \\

MultiVI
& 0.093 & 0.007 & 0.834 & 0.784 & 0.753 & 0.830
& 0.061 & 0.011 & 0.777 & 0.642 & 0.585 & 0.782 \\

Cobolt
& 0.088 & 0.012 & 0.812 & 0.799 & 0.786 & 0.921
& 0.097 & 0.006 & 0.742 & 0.740 & 0.512 & 0.946 \\

\midrule
\multicolumn{13}{l}{\textit{Paired target-space mappers}} \\

MSE
& 0.048 & 0.110 & 0.902 & 0.821 & 0.814 & 0.973
& 0.042 & 0.032 & 0.786 & 0.757 & 0.711 & 0.980 \\

MSE+MMD
& 0.047 & 0.120 & 0.917 & 0.768 & 0.811 & 0.982
& 0.039 & 0.034 & 0.791 & 0.769 & 0.726 & 0.987 \\

MSE+InfoNCE
& 0.053 & 0.115 & 0.914 & 0.823 & 0.817 & 0.971
& 0.040 & 0.034 & 0.774 & 0.751 & 0.700 & 0.983 \\

MSE+Neighborhood
& 0.047 & 0.119 & 0.898 & 0.838 & 0.845 & 0.981
& 0.041 & 0.038 & 0.796 & 0.766 & 0.726 & \textbf{0.992} \\
\midrule
FlowAlign
& \textbf{0.039} & \textbf{0.141} & \textbf{0.920} & \textbf{0.872} & \textbf{0.885} & 0.978
& \textbf{0.019} & \textbf{0.050} & \textbf{0.836} & \textbf{0.797} & \textbf{0.768} & 0.988 \\

\bottomrule
\end{tabular}
}
\end{table}

\begin{figure}[t]
  \centering
  \includegraphics[width=1\textwidth]{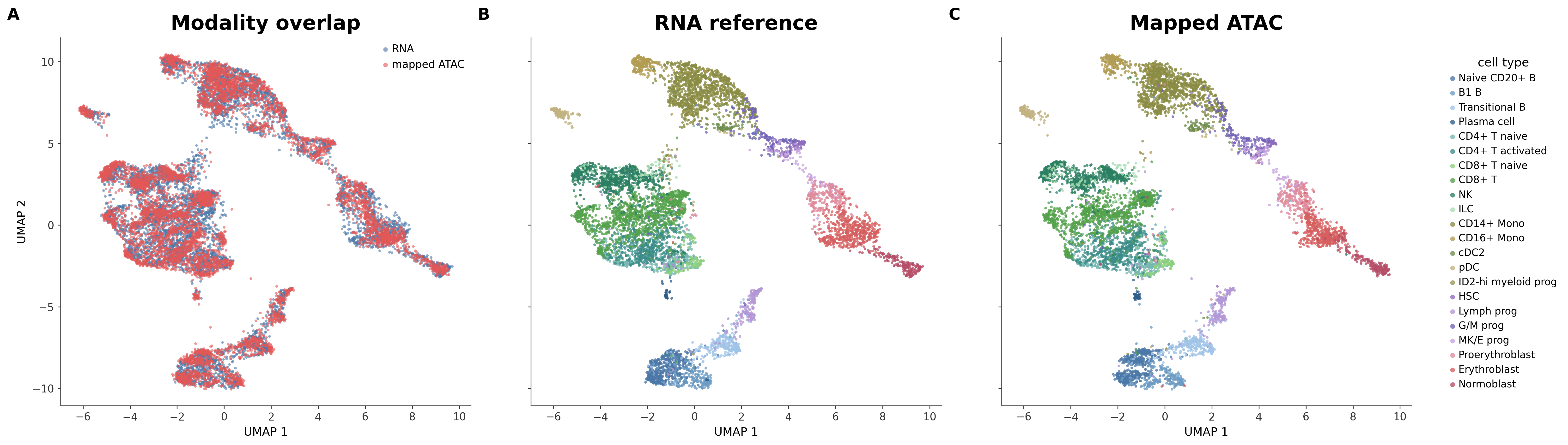}
  \caption{BMMC held-out target-space visualization.
(A) UMAP overlap between held-out RNA embeddings and mapped ATAC embeddings.
(B) RNA reference embeddings colored by cell type.
(C) Mapped ATAC embeddings colored by the same cell-type labels.}
  \label{fig:test_umap}
\end{figure}

\paragraph{FlowAlign as a Regularizer.}
We also validate that FlowAlign can be used as a plug-in regularizer on the PBMC multiome data. 
We first train scVI~\citep{lopez2018deep} and PeakVI~\citep{ashuach2022peakvi}, variational autoencoder-based models for scRNA-seq and scATAC-seq data, respectively. 
These models take raw modality-specific count data as input and learn latent representations with decoders for reconstruction. 
The scVI RNA latent space is kept fixed as the target space, and FlowAlign is used to reshape the PeakVI ATAC latent space toward this target. 
Starting from the pretrained PeakVI model, we fine-tune it with PeakVI peak reconstruction loss and the FlowAlign path-consistency loss. 
The results show that the regularized PeakVI latent achieves alignment performance comparable to strong baselines, allowing ATAC and RNA cells to be analyzed in the same latent coordinate system, while largely preserving PeakVI's peak reconstruction ability. 
This suggests that FlowAlign can serve as a regularizer for models with existing task-specific objectives, improving alignment to a fixed target space without substantially sacrificing the model's original task performance. Detailed results are provided in Appendix~\ref{app:peakvi_regularizer}.

\subsection{Image--Text Target-Space Alignment}
\label{sec:exp_image_text}
To evaluate whether FlowAlign generalizes beyond single-cell data, we conduct an additional experiment on CUBICC~\citep{palumbo2024deep}, a dataset containing paired bird images and textual descriptions.
We extract image representations using DINOv3~\citep{simeoni2025dinov3} and text representations using SigLIP2~\citep{tschannen2025siglip}.
Since the two encoders produce embeddings of different dimensions, we independently apply PCA to obtain 128-dimensional representations for both modalities.
PCA is fitted only on the training embeddings, and the resulting transformations are applied unchanged to the validation and test sets.

The text representations define the fixed target space, and FlowAlign learns to map image representations into this space using paired image--text supervision.
At inference time, only image representations are provided.
We evaluate paired correspondence using FOSCTTM and exact pair retrieval (R@1), together with label transfer accuracy (LTA).
We compare FlowAlign with direct MSE regression and MSE-based regularization using MMD, InfoNCE, and neighborhood preservation.

\begin{table}[t]
\centering
\caption{
Performance comparison on paired image--text alignment using CUBICC.
All methods map DINOv3 image representations into the fixed 128-dimensional SigLIP2 text space.
}
\label{tab:cubicc}
\small
\begin{tabular}{lccc}
\toprule
\textbf{Method}
& \textbf{FOSCTTM} $\downarrow$
& \textbf{R@1} $\uparrow$
& \textbf{LTA (\%)} $\uparrow$ \\
\midrule
Identity
& 0.499 & 0.003 & 13.31 \\
MSE
& 0.069 & 0.067 & 57.49 \\
MSE+MMD
& 0.076 & 0.052 & 63.95 \\
MSE+InfoNCE
& 0.078 & 0.051 & 59.77 \\
MSE+Neighborhood
& 0.074 & 0.063 & 68.90 \\
\midrule
FlowAlign
& \textbf{0.063} & \textbf{0.075} & \textbf{70.42} \\
\bottomrule
\end{tabular}
\end{table}

As shown in Table~\ref{tab:cubicc}, FlowAlign achieves the best performance across all three metrics.
The results provide additional evidence that the path-consistency objective is not restricted to single-cell modalities, but can also be applied to image--text alignment in a fixed target representation space.

\subsection{Ablation studies}
\label{sec:exp_ablation}

We perform ablations on PBMC to isolate the main design choices of the proposed path-consistency score. 
All variants use the same paired-training and source-only inference protocol as the full method.

Table~\ref{tab:ablation_pbmc} summarizes the component ablations. 
Removing local-start sampling by setting \(\sigma=0\) substantially degrades exact cell matching and structure-aware metrics, showing that a local neighborhood of paths is more informative than a single deterministic segment from \(s\) to \(t\). 
Removing source conditioning from \(v_\psi(x_\tau,\tau;s)\) also reduces performance, indicating that the source context provides instance-specific path information beyond the source-dependent start distribution. 
CFM transport directly follows the learned velocity field as an inference-time transport map. While it remains competitive on pair-level retrieval metrics, its lower LTA, NMI, and ARI indicate weaker preservation of biological structure. 
This suggests that the learned field is more useful here as a training-time path-consistency score than as a direct test-time transport map.
Finally, the degenerate situation removes the candidate path-location input \(x_\tau\), reducing the score to an MSE-like source-conditioned velocity teacher. 
Its gap from the full model supports the value of evaluating the frozen field along candidate-dependent paths.

\begin{wrapfigure}{r}{0.5\linewidth}
  \centering
  \vspace{-10pt}
  \includegraphics[width=0.50\textwidth]{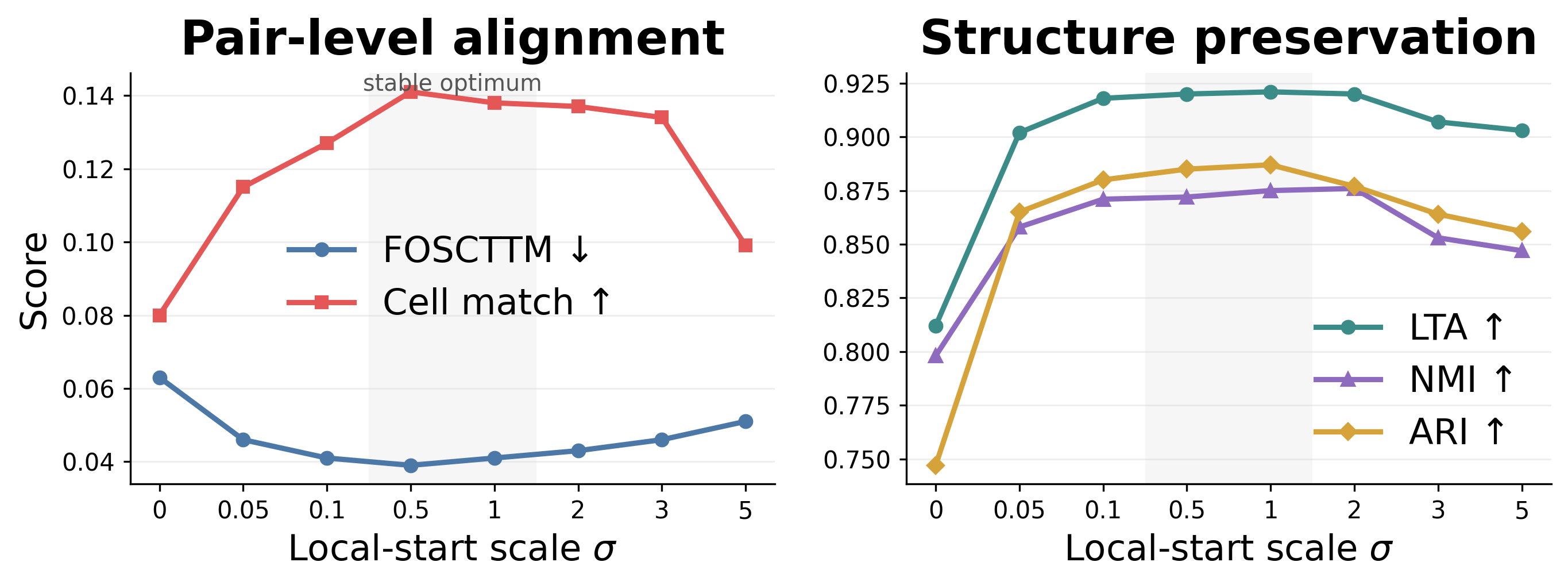}
  \caption{
  Effect of the local-start scale \(\sigma\) on PBMC.
  }
  \label{fig:sigma_sweep}
  \vspace{-10pt}
\end{wrapfigure}

Figure~\ref{fig:sigma_sweep} further examines the local-start scale \(\sigma\). 
We use the median 10-nearest-neighbor distance in the standardized RNA PCA target space, \(3.14\) on PBMC, as a reference scale for local target-space geometry. 
Moving from \(\sigma=0\) to a moderate local-start radius improves both pair-level alignment and structure preservation, while very large \(\sigma\) begins to degrade performance. 
This supports the interpretation that \(p_0(\cdot\mid s)\) should define a local tube around the source-conditioned start rather than an overly diffuse set of paths. Additional Stage-1 field diagnostics on held-out data are provided in Appendix~\ref{app:field_diagnostics}.

\begin{table}[t]
\centering
\caption{
Ablation study results on PBMC dataset.
}
\label{tab:ablation_pbmc}
\resizebox{0.8\textwidth}{!}{
\begin{tabular}{lccccc}
\toprule
\textbf{Variant} 
& \textbf{FOSCTTM} $\downarrow$
& \textbf{Cell match} $\uparrow$
& \textbf{LTA} $\uparrow$
& \textbf{NMI} $\uparrow$
& \textbf{ARI} $\uparrow$ \\
\midrule
FlowAlign 
& 0.039 & 0.141 & \textbf{0.920} & \textbf{0.872} & \textbf{0.885} \\
$\sigma=0$ 
& 0.063 & 0.082 & 0.812 & 0.798 & 0.747 \\
Source-free field
& 0.055 & 0.104 & 0.872 & 0.809 & 0.822 \\
CFM transport
& \textbf{0.034} & \textbf{0.142} & 0.892 & 0.826 & 0.834 \\
Degenerate case
& 0.047 & 0.115 & 0.902 & 0.857 & 0.865 \\
\bottomrule
\end{tabular}
}
\end{table}

\section{Conclusion, Limitations, and Future Work}


We presented FlowAlign, a flow-guided method for fixed target-space alignment based on a path-consistency score.
FlowAlign trains a source-conditioned velocity field along paired interpolation paths, freezes it, and uses its velocity residual as a training-time compatibility signal rather than a test-time transport mechanism.
Our analysis shows that the path-consistency loss can control endpoint error when the frozen field is accurate on paired paths and that, locally, it behaves as a learned source-conditioned metric rather than a fixed Euclidean penalty.
Experiments on synthetic data, paired single-cell multiome data, and image--text alignment show that the method improves instance-level correspondence while preserving target-space structure.
A limitation is that FlowAlign requires paired data, a meaningful common vector space for path construction, and an accurate Stage-1 field; incompatible source representations or poorly learned fields may weaken the alignment signal.
Moreover, the current analysis assumes Euclidean linear paths, and standalone alignment does not guarantee preservation of source-specific information absent from the target space.
Future work will explore more flexible start distributions, field diagnostics, partially paired or unpaired settings, and broader applications.

\newpage

\begin{ack}
This work was supported in part by the National Institutes of Health (NIH) under grant R56LM014522 and the National Science Foundation (NSF) under grant DMS-2310217. This research was supported in part by the University of Pittsburgh Center for Research Computing and Data, RRID:SCR\_022735, through the resources provided.

\textbf{Competing Interests.}
The authors declare no competing interests.
\end{ack}
{\small
\bibliographystyle{plain}
\bibliography{references}
}







\newpage
\appendix



\section{Additional Theory}
\label{app:theory}

\subsection{Proof of Proposition~\ref{prop:endpoint_control}}
\label{app:proof_endpoint_control}

We prove the approximate endpoint-control bound stated in Proposition~\ref{prop:endpoint_control}. 
Fix a paired example \((s,t)\) and a candidate endpoint \(\tilde t\). 
For \(x_0\sim p_0(\cdot\mid s)\), define the paired and candidate paths as
\(x_\tau^t=(1-\tau)x_0+\tau t\) and 
\(x_\tau^{\tilde t}=(1-\tau)x_0+\tau \tilde t\). 
Let \(\delta=\tilde t-t\). 
Then \(x_\tau^{\tilde t}-x_\tau^t=\tau\delta\).

Define the paired-path residual
\[
r_t(x_0,\tau;s)
=
v_\psi(x_\tau^t,\tau;s)-(t-x_0),
\]
so that \(\varepsilon_\psi(s,t)^2=\mathbb E_{x_0,\tau}\|r_t(x_0,\tau;s)\|_2^2\). 
The candidate residual can be decomposed as
\[
v_\psi(x_\tau^{\tilde t},\tau;s)-(\tilde t-x_0)
=
\big[
v_\psi(x_\tau^{\tilde t},\tau;s)-v_\psi(x_\tau^t,\tau;s)
\big]
-\delta
+
r_t(x_0,\tau;s).
\]
Let 
\[
a(x_0,\tau)
=
\big[
v_\psi(x_\tau^{\tilde t},\tau;s)-v_\psi(x_\tau^t,\tau;s)
\big]
-\delta .
\]
Then
\[
\sqrt{\mathcal C_\psi(\tilde t;s)}
=
\|a+r_t\|_{L^2(x_0,\tau)}
\ge
\|a\|_{L^2(x_0,\tau)}
-
\|r_t\|_{L^2(x_0,\tau)}
=
\|a\|_{L^2(x_0,\tau)}
-
\varepsilon_\psi(s,t).
\]

It remains to lower-bound \(\|a\|_{L^2}\). 
By the reverse triangle inequality and the \(L\)-Lipschitz assumption,
\[
\begin{aligned}
\|a(x_0,\tau)\|_2
&=
\left\|
\big[
v_\psi(x_\tau^{\tilde t},\tau;s)-v_\psi(x_\tau^t,\tau;s)
\big]
-\delta
\right\|_2 \\
&\ge
\|\delta\|_2
-
\left\|
v_\psi(x_\tau^{\tilde t},\tau;s)-v_\psi(x_\tau^t,\tau;s)
\right\|_2 \\
&\ge
(1-L\tau)\|\delta\|_2 .
\end{aligned}
\]
Since the left-hand side is nonnegative, we have
\[
\|a(x_0,\tau)\|_2
\ge
(1-L\tau)_+\|\delta\|_2 .
\]
Taking the \(L^2(x_0,\tau)\) norm gives
\[
\|a\|_{L^2(x_0,\tau)}
\ge
\left(\int_0^1(1-L\tau)_+^2\,d\tau\right)^{1/2}
\|\delta\|_2
=
\sqrt{\kappa(L)}\|\tilde t-t\|_2 .
\]
Combining the inequalities yields
\[
\sqrt{\mathcal C_\psi(\tilde t;s)}
\ge
\sqrt{\kappa(L)}\|\tilde t-t\|_2
-
\varepsilon_\psi(s,t),
\]
or equivalently,
\[
\|\tilde t-t\|_2
\le
\frac{
\sqrt{\mathcal C_\psi(\tilde t;s)}
+
\varepsilon_\psi(s,t)
}{
\sqrt{\kappa(L)}
}.
\]
This proves the claim.

For reference, \(\kappa(L)\) has the closed form
\[
\kappa(L)
=
\begin{cases}
1-L+\dfrac{L^2}{3}, & 0\le L\le 1,\\[6pt]
\dfrac{1}{3L}, & L>1.
\end{cases}
\]
Therefore \(\kappa(L)>0\) for any finite \(L\).

\subsection{Path-Level Residual Decomposition}
\label{app:path_residual_decomposition}

We first record the basic decomposition behind the path-consistency critic. 
Fix a paired example \((s,t)\) and a candidate endpoint \(\tilde t\). 
For the same start point \(x_0\), define
\(x_\tau^t=(1-\tau)x_0+\tau t\) and 
\(x_\tau^{\tilde t}=(1-\tau)x_0+\tau\tilde t\). 
Let \(\delta=\tilde t-t\). 
Then
\[
x_\tau^{\tilde t}-x_\tau^t=\tau\delta .
\]

The candidate residual is
\[
r_{\tilde t}(x_0,\tau;s)
=
v_\psi(x_\tau^{\tilde t},\tau;s)-(\tilde t-x_0),
\]
while the paired-path residual is
\[
r_t(x_0,\tau;s)
=
v_\psi(x_\tau^t,\tau;s)-(t-x_0).
\]
Adding and subtracting \(v_\psi(x_\tau^t,\tau;s)\), we obtain
\[
\begin{aligned}
r_{\tilde t}(x_0,\tau;s)
&=
v_\psi(x_\tau^{\tilde t},\tau;s)-(\tilde t-x_0)\\
&=
\left[
v_\psi(x_\tau^{\tilde t},\tau;s)
-
v_\psi(x_\tau^t,\tau;s)
\right]
-
(\tilde t-t)
+
r_t(x_0,\tau;s).
\end{aligned}
\]
In the realizable case where \(r_t(x_0,\tau;s)=0\), this reduces to
\[
r_{\tilde t}(x_0,\tau;s)
=
\left[
v_\psi(x_\tau^{\tilde t},\tau;s)
-
v_\psi(x_\tau^t,\tau;s)
\right]
-
\delta .
\]
Thus, the candidate endpoint error enters the score through both the required path velocity \(\delta\) and the change in the frozen field along the perturbed path.

\subsection{Average Endpoint Error Bound}
\label{app:average_endpoint_bound}

Let \(\hat t=g_\theta(s)\). 
From Proposition~\ref{prop:endpoint_control},
\[
\|\hat t-t\|_2
\le
\frac{
\sqrt{\mathcal C_\psi(g_\theta(s);s)}
+
\varepsilon_\psi(s,t)
}{
\sqrt{\kappa(L)}
}.
\]
Squaring both sides and using \((a+b)^2\le 2a^2+2b^2\), we obtain
\[
\|\hat t-t\|_2^2
\le
\frac{2}{\kappa(L)}
\left(
\mathcal C_\psi(g_\theta(s);s)
+
\varepsilon_\psi(s,t)^2
\right).
\]
Taking expectation over \(s\) gives
\[
\mathbb E_s\|\hat t-t\|_2^2
\le
\frac{2}{\kappa(L)}
\left(
\mathbb E_s[\mathcal C_\psi(g_\theta(s);s)]
+
\mathbb E_s[\varepsilon_\psi(s,t)^2]
\right).
\]
This is the average endpoint error bound used in the main text.

\subsection{Local Quadratic Approximation}
\label{app:local_quadratic}

We derive the local metric interpretation of the path-consistency critic. 
Let \(\delta=\tilde t-t\). 
The paired and candidate paths satisfy \(x_\tau^{\tilde t}=x_\tau^t+\tau\delta\). 
Assume \(v_\psi\) is differentiable in its state argument near the paired path. 
A first-order Taylor expansion gives
\[
v_\psi(x_\tau^{\tilde t},\tau;s)
=
v_\psi(x_\tau^t,\tau;s)
+
\tau J_\psi(x_\tau^t,\tau;s)\delta
+
o(\|\delta\|_2),
\]
where \(J_\psi(x,\tau;s)=\nabla_x v_\psi(x,\tau;s)\). 
If the paired-path residual is negligible, \(v_\psi(x_\tau^t,\tau;s)\approx t-x_0\). 
Therefore,
\[
v_\psi(x_\tau^{\tilde t},\tau;s)-(\tilde t-x_0)
\approx
(\tau J_\psi(x_\tau^t,\tau;s)-I)\delta .
\]
Substituting this expression into the definition of \(\mathcal C_\psi\) yields
\[
\mathcal C_\psi(\tilde t;s)
\approx
\delta^\top A_\psi(s)\delta,
\]
where
\[
A_\psi(s)
=
\mathbb E_{x_0,\tau}
\left[
(I-\tau J_\psi(x_\tau^t,\tau;s))^\top
(I-\tau J_\psi(x_\tau^t,\tau;s))
\right].
\]
Thus, locally, the critic penalizes endpoint errors through a field-induced quadratic form rather than the fixed identity metric used by endpoint MSE.

\subsection{Source Conditioning Reduces Velocity Ambiguity}
\label{app:source_conditioning}

We justify the role of source conditioning in the velocity field. 
Let \(X_\tau\) denote a path point and \(V\) denote the corresponding target velocity. 
For squared velocity regression, the optimal unconditioned predictor at \((X_\tau,\tau)\) is
\[
v^\star(X_\tau,\tau)
=
\mathbb E[V\mid X_\tau,\tau].
\]
The corresponding irreducible conditional risk is
\[
\inf_v
\mathbb E[\|V-v\|_2^2\mid X_\tau,\tau]
=
\mathrm{Tr}(\mathrm{Var}(V\mid X_\tau,\tau)).
\]

If the field is conditioned on the source representation \(S\), the optimal predictor becomes
\[
v^\star(X_\tau,\tau;S)
=
\mathbb E[V\mid X_\tau,\tau,S],
\]
with irreducible risk
\[
\mathrm{Tr}(\mathrm{Var}(V\mid X_\tau,\tau,S)).
\]
By the law of total variance,
\[
\mathrm{Var}(V\mid X_\tau,\tau)
=
\mathbb E[
\mathrm{Var}(V\mid X_\tau,\tau,S)
\mid X_\tau,\tau]
+
\mathrm{Var}(
\mathbb E[V\mid X_\tau,\tau,S]
\mid X_\tau,\tau).
\]
Taking traces and expectations gives
\[
\mathbb E\!\left[
\mathrm{Tr}(\mathrm{Var}(V\mid X_\tau,\tau,S))
\right]
\le
\mathbb E\!\left[
\mathrm{Tr}(\mathrm{Var}(V\mid X_\tau,\tau))
\right].
\]
The inequality is strict whenever \(S\) explains velocity variation not determined by \((X_\tau,\tau)\). 
Thus, source conditioning can reduce the irreducible error caused by intersecting or nearby paths that require different target velocities.

\subsection{Neighborhood Stability from Endpoint Control}
\label{app:neighborhood_stability}

Endpoint control also implies a simple stability guarantee for local distances. 
Suppose predictions \(\hat t_i\) satisfy \(\|\hat t_i-t_i\|_2\le \varepsilon\) for all \(i\). 
Then, for any pair \(i,j\),
\[
\begin{aligned}
\left|
\|\hat t_i-\hat t_j\|_2-\|t_i-t_j\|_2
\right|
&\le
\|(\hat t_i-\hat t_j)-(t_i-t_j)\|_2 \\
&\le
\|\hat t_i-t_i\|_2+\|\hat t_j-t_j\|_2 \\
&\le
2\varepsilon .
\end{aligned}
\]
Thus, small endpoint errors imply small additive distortion of pairwise distances in the target space.

This also yields a standard nearest-neighbor stability condition. 
For a point \(i\), let \(d_{i,(k)}\) and \(d_{i,(k+1)}\) be the \(k\)-th and \((k+1)\)-th nearest-neighbor distances in the true target space, and let \(\gamma_i=d_{i,(k+1)}-d_{i,(k)}\). 
If \(4\varepsilon<\gamma_i\), then the top-\(k\) neighbor set of \(i\) is preserved under the predictions. 
Indeed, distances to true top-\(k\) neighbors can increase by at most \(2\varepsilon\), while distances to non-neighbors can decrease by at most \(2\varepsilon\). 
The margin condition prevents a non-neighbor from overtaking a true top-\(k\) neighbor.

\paragraph{Nearest-neighbor label stability.}
A similar argument gives a label-stability condition. 
Let \(\ell_i\) denote the label of sample \(i\). 
For a fixed point \(i\), define the nearest same-label and different-label target distances
\[
a_i=\min_{j:\ell_j=\ell_i,\ j\neq i}\|t_i-t_j\|_2,
\qquad
b_i=\min_{j:\ell_j\neq \ell_i}\|t_i-t_j\|_2,
\]
and let \(m_i=b_i-a_i\) be the label margin. 
If \(\|\hat t_j-t_j\|_2\le\varepsilon\) for all \(j\) and \(4\varepsilon<m_i\), then the nearest neighbor of \(\hat t_i\) in the predicted space has the same label as \(i\).

To see this, the predicted distance from \(i\) to its nearest same-label target neighbor is at most \(a_i+2\varepsilon\), while the predicted distance to any different-label point is at least \(b_i-2\varepsilon\). 
The condition \(a_i+2\varepsilon<b_i-2\varepsilon\), equivalently \(4\varepsilon<m_i\), guarantees that a same-label neighbor remains closer than every different-label point.

\newpage
\section{Experiment Implementation Details}
\label{app:experiment_details}

This appendix summarizes the network architectures and main training hyperparameters used in the synthetic toy, PBMC, and BMMC experiments. All experiments are conducted using Nvidia H100 GPU with 80GB memory.

\subsection{PBMC and BMMC Experiments}
\label{app:pbmc_bmmc_arch}
All data are split into 80\% training samples, 10\% validation samples and 10\% testing samples.
\paragraph{Stage-1 velocity field.}
For both PBMC and BMMC, the Stage-1 velocity field is a source-conditioned residual MLP. 
The input is the concatenation of the bridge location, the ATAC latent condition, and a Fourier time embedding:
\[
25 + 25 + 64 = 114.
\]
The output is a 25-dimensional velocity in the RNA latent space. 
The architecture is:
\[
\begin{array}{l}
\mathrm{Linear}(114,128) \rightarrow \mathrm{SiLU} \\
6 \times \mathrm{ResidualFlowBlock}(128) \\
\mathrm{LayerNorm} \rightarrow \mathrm{Linear}(128,128) \rightarrow \mathrm{SiLU} \\
\mathrm{Linear}(128,25)
\end{array}
\]
Each residual flow block contains LayerNorm, two linear layers, a SiLU activation, and a residual connection.

\paragraph{Stage-2 mapper.}
The Stage-2 mapper maps ATAC latent representations to RNA latent representations. 
For both PBMC and BMMC, the architecture is:

\[
\begin{array}{l}
\mathrm{Linear}(25,512) \rightarrow \mathrm{SiLU} \\
6 \times \mathrm{ResidualFlowBlock}(512) \\
\mathrm{LayerNorm} \rightarrow \mathrm{Linear}(512,512) \rightarrow \mathrm{SiLU} \\
\mathrm{Linear}(512,25)
\end{array}
\]

Each residual block contains LayerNorm, two linear layers, a SiLU activation, dropout with rate 0, and a residual connection.

\paragraph{Hyperparameters.}
The main hyperparameters are summarized in Table~\ref{tab:pbmc_bmmc_hyperparams}.

\begin{table}[h]
\centering
\caption{Network and training hyperparameters for PBMC and BMMC experiments.}
\label{tab:pbmc_bmmc_hyperparams}
\small
\begin{tabular}{lcc}
\toprule
\textbf{Setting} & \textbf{PBMC} & \textbf{BMMC} \\
\midrule
Latent dimension & 25 & 25 \\
Stage-1 input dimension & 114 & 114 \\
Stage-1 hidden dimension & 128 & 128 \\
Stage-1 residual blocks & 6 & 6 \\
Stage-1 batch size & 256 & 256 \\
Stage-1 learning rate & \(3\times10^{-4}\) & \(3\times10^{-4}\) \\
Stage-1 weight decay & \(10^{-4}\) & \(10^{-4}\) \\
\midrule
Stage-2 input dimension & 25 & 25 \\
Stage-2 hidden dimension & 512 & 512 \\
Stage-2 residual blocks & 6 & 6 \\
Stage-2 batch size & 256 & 256 \\
Stage-2 max epochs & 1000 & 1000 \\
Stage-2 learning rate & \(2\times10^{-4}\) & \(2\times10^{-4}\) \\
Stage-2 weight decay & \(3\times10^{-4}\) & \(3\times10^{-4}\) \\
\bottomrule
\end{tabular}
\end{table}

\subsection{Synthetic Toy Experiment}
\label{app:toy_arch}

\paragraph{Generic MLP.}
All toy networks use SiLU MLPs. 
The generic MLP has the form:
\[
\mathrm{Linear}(d_{\mathrm{in}}, h)
\rightarrow
\mathrm{SiLU}
\rightarrow
\cdots
\rightarrow
\mathrm{Linear}(h,h)
\rightarrow
\mathrm{SiLU}
\rightarrow
\mathrm{Linear}(h,d_{\mathrm{out}}).
\]
Unless otherwise specified, the hidden dimension is 128 and the depth is 3. 
The generic toy MLP does not use LayerNorm, batch normalization, dropout, or residual blocks.

\paragraph{Hyperparameters.}
All toy networks are trained with AdamW, learning rate \(10^{-3}\), batch size 256, epochs 1000, weight decay \(10^{-4}\). 

\subsection{Evaluation metrics}
\label{app:evaluation_metrics}

We evaluate single-cell multiome alignment using two groups of metrics: 
paired-cell correspondence metrics and biological-structure metrics. 
All metrics are computed on held-out test cells. 
For our method, distances are computed between mapped ATAC embeddings and fixed RNA PCA target embeddings. 
For baseline methods with native aligned spaces, distances are computed in the corresponding aligned embedding spaces using the same held-out paired-cell protocol.

\paragraph{Paired-cell correspondence.}
Let \(\hat t_i\) denote the mapped ATAC embedding of test cell \(i\), and let \(t_i\) denote its paired RNA embedding. 
For each mapped ATAC cell \(\hat t_i\), we rank all RNA test-cell embeddings \(\{t_j\}_{j=1}^n\) by Euclidean distance to \(\hat t_i\). 
We report FOSCTTM~\cite{liu2019jointly} and exact cell match.

FOSCTTM measures the fraction of RNA cells that are closer to \(\hat t_i\) than the true paired RNA cell \(t_i\):
\[
\mathrm{FOSCTTM}
=
\frac{1}{n}
\sum_{i=1}^n
\frac{
\sum_{j\neq i}
\mathbf 1\{
\|\hat t_i-t_j\|_2 < \|\hat t_i-t_i\|_2
\}
}{n-1}.
\]
Lower FOSCTTM indicates better paired-cell recovery.

Exact cell match, or R@1, measures whether the true paired RNA cell is the nearest neighbor of the mapped ATAC cell:
\[
\mathrm{R@1}
=
\frac{1}{n}
\sum_{i=1}^n
\mathbf 1
\left\{
i =
\arg\min_j
\|\hat t_i-t_j\|_2
\right\}.
\]
Higher R@1 indicates stronger instance-level correspondence.

\paragraph{Biological structure.}
We evaluate preservation of biological organization using label transfer accuracy (LTA), normalized mutual information (NMI), adjusted Rand index (ARI), and omics mixing based on average silhouette width (ASW-O)~\cite{xiao2024benchmarking}.

For LTA, each mapped ATAC cell is assigned the cell-type label of its nearest RNA neighbor in the embedding space. 
The transferred label is then compared with the ground-truth cell-type label of the same test cell:
\[
\mathrm{LTA}
=
\frac{1}{n}
\sum_{i=1}^n
\mathbf 1
\left\{
y_i =
y_{\arg\min_j \|\hat t_i-t_j\|_2}
\right\},
\]
where \(y_i\) is the annotated cell type of cell \(i\). 
Higher LTA indicates better preservation of RNA-space cell-type neighborhoods.

For NMI and ARI, we cluster the combined set of mapped ATAC embeddings and RNA embeddings using the same clustering procedure for all methods. 
We then compare the resulting cluster assignments with annotated cell-type labels. 
NMI measures the mutual information between clusters and labels after normalization, while ARI measures pairwise agreement between clusters and labels after correcting for chance. 
Higher NMI and ARI indicate stronger preservation of biological cell-type structure.

ASW-O evaluates whether embeddings from different omics modalities are well mixed. 
We compute silhouette widths using modality labels, treating mapped ATAC and RNA as two modalities, and report the transformed ASW-O score following the benchmark convention, where larger values indicate better cross-omics mixing. 
Thus, high ASW-O means that mapped ATAC cells and RNA cells with similar biological states are not strongly separated by modality.

\subsection{Additional Experimental Details}

For the controlled PBMC and BMMC comparisons, all MSE-based baselines use the same mapper architecture, data splits, and training protocol as FlowAlign.
The neighborhood-preservation baseline penalizes discrepancies in pairwise distances along target-defined local neighbor edges.

For CUBICC, we extract paired image and text embeddings using DINOv3 and SigLIP2, respectively.
Each modality is independently reduced to 128 dimensions using PCA fitted on training data only.
The learned PCA transformations are applied unchanged to validation and test sets, with text embeddings defining the fixed target space.

\paragraph{Computational cost.}
On a single NVIDIA H100, FlowAlign takes approximately 5.5 and 39.5 minutes to train on PBMC and BMMC, respectively, compared with 2.1 and 12.0 minutes for the architecture-matched MSE mapper.
These timings exclude preprocessing and evaluation.

\section{Additional Results}
\subsection{Additional Toy Experiment Results}
\label{app:toy_exp}
\begin{figure}[h]
  \centering
  \includegraphics[width=0.6\textwidth]{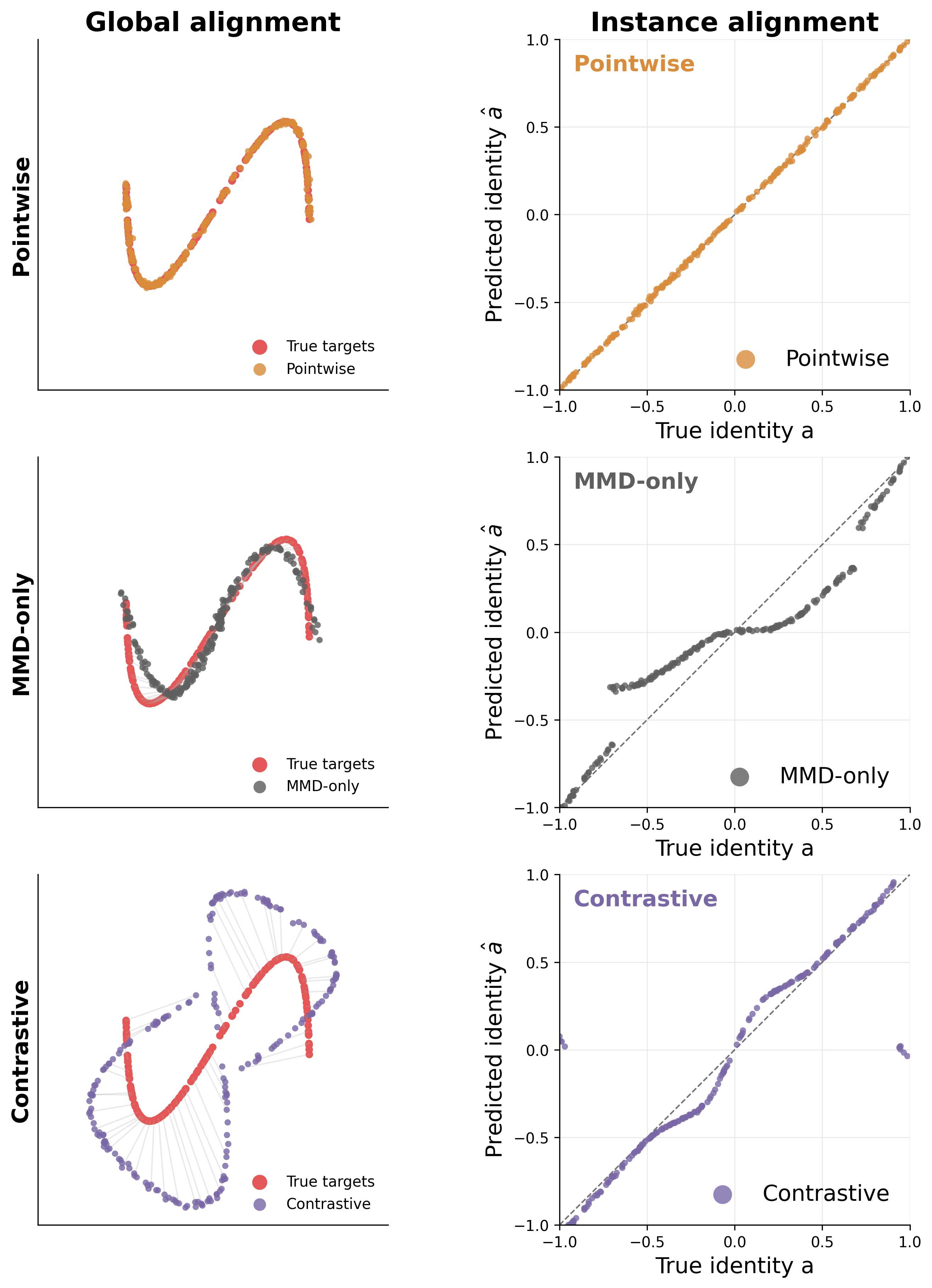}
  \caption{Visualization of baseline results of synthetic fixed-target alignment.
}
  \label{fig:toy_supp}
\end{figure}

Figure~\ref{fig:toy_supp} provides additional visualizations for the synthetic fixed-target alignment experiment. 
These plots complement the quantitative results in the main text by showing how different objectives affect the geometry of the mapped source points in the fixed target space.

\subsection{Additional Single Cell Experiment Results}
\label{app:sc_supp}
We provide additional single-cell alignment metrics to complement the main results. 
The main text reports a compact set of paired-cell correspondence and biological-structure metrics. 
Here, we include supplementary retrieval metrics and additional cell-type structure metrics on the same held-out PBMC and BMMC test splits.

\paragraph{UMAP visualization of the PBMC experiment.}

\begin{figure}[h]
  \centering
  \includegraphics[width=1\textwidth]{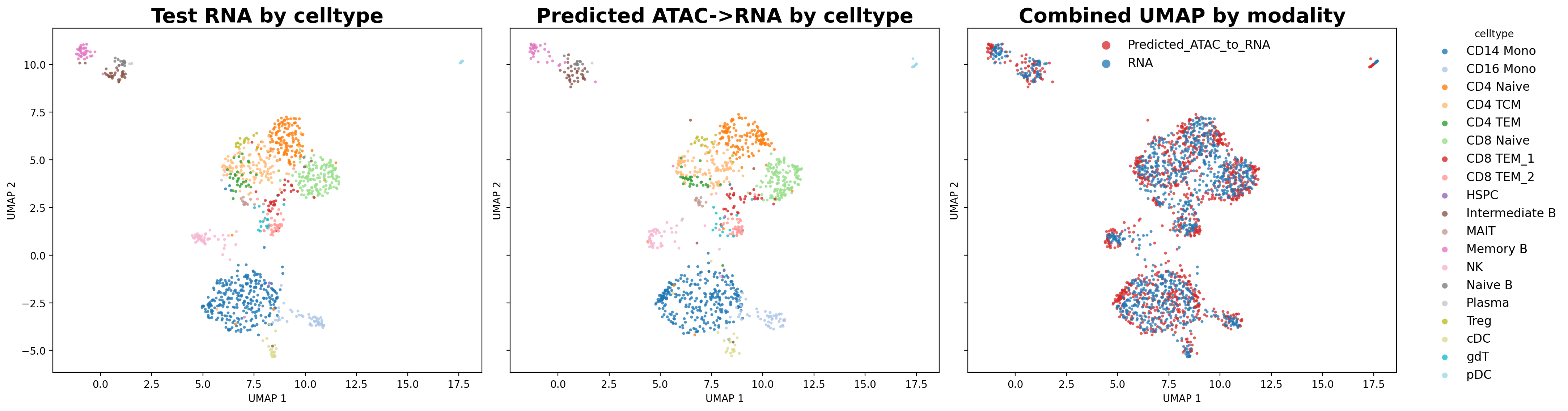}
  \caption{PBMC held-out target-space visualization.
(A) RNA reference embeddings colored by cell type.
(B) Mapped ATAC embeddings colored by the same cell-type labels.
(C) UMAP overlap between held-out RNA embeddings and mapped ATAC embeddings}
  \label{fig:pbmc_umap}
\end{figure}

Figure~\ref{fig:pbmc_umap} shows the PBMC UMAP visualization of held-out RNA embeddings and mapped ATAC embeddings. 
The mapped ATAC cells overlap the RNA reference geometry, providing a qualitative check of target-space structure preservation.

\paragraph{Additional paired-cell retrieval metrics.}

Table~\ref{tab:additional_retrieval_metrics} reports retrieval-based metrics beyond the main FOSCTTM and R@1 results. 
Median FOSCTTM summarizes the typical paired-cell retrieval error, while MRR and recall at larger cutoffs measure whether the true paired RNA cell appears near the top of the ranked list. 
FlowAlign improves these retrieval metrics on both PBMC and BMMC, indicating that the mapped ATAC embeddings better preserve fine-grained paired-cell correspondence in the target space.

\begin{table}[h]
\centering
\caption{
Additional paired-cell retrieval metrics on PBMC and BMMC.
Lower FOSCTTM median and mean rank are better; higher MRR and recall values are better.
}
\label{tab:additional_retrieval_metrics}
\resizebox{\textwidth}{!}{
\begin{tabular}{lcccccccccc}
\toprule
\multirow{2}{*}{\textbf{Method}}
& \multicolumn{5}{c}{\textbf{PBMC}}
& \multicolumn{5}{c}{\textbf{BMMC}} \\
\cmidrule(lr){2-6}
\cmidrule(lr){7-11}
& \textbf{FOSCTTM med.} $\downarrow$
& \textbf{MRR} $\uparrow$
& \textbf{R@5} $\uparrow$
& \textbf{R@10} $\uparrow$
& \textbf{R@50} $\uparrow$
& \textbf{FOSCTTM med.} $\downarrow$
& \textbf{MRR} $\uparrow$
& \textbf{R@5} $\uparrow$
& \textbf{R@10} $\uparrow$
& \textbf{R@50} $\uparrow$ \\
\midrule
Seurat v5
& 0.026 & 0.079 & 0.096 & 0.185 & 0.564
& 0.020 & 0.032 & 0.030 & 0.064 & 0.250 \\

GLUE
& 0.023 & 0.139 & 0.155 & 0.247 & 0.621
& 0.024 & 0.046 & 0.027 & 0.059 & 0.312 \\

MultiVI
& 0.053 & 0.043 & 0.042 & 0.100 & 0.279
& 0.034 & 0.028 & 0.021 & 0.056 & 0.196 \\

Cobolt
& 0.053 & 0.050 & 0.061 & 0.156 & 0.555
& 0.042 & 0.029 & 0.017 & 0.041 & 0.242 \\

FlowAlign
& 0.011 & 0.241 & 0.321 & 0.464 & 0.797
& 0.006 & 0.116 & 0.158 & 0.237 & 0.528 \\

\bottomrule
\end{tabular}
}
\end{table}

\paragraph{Additional biological-structure metrics.}
Table~\ref{tab:additional_bio_metrics} reports supplementary biological-structure metrics based on fine cell-type labels. 
Balanced LTA and Macro-F1 account for class imbalance across cell types, while mAP evaluates the quality of the full same-label retrieval ranking. 
FlowAlign obtains the best Balanced LTA and Macro-F1 on both datasets, supporting the conclusion that the mapped ATAC embeddings preserve RNA-space cell-type neighborhoods. 
Although mAP is more sensitive to the global ordering of all same-label cells, the results remain competitive and complement the local label-transfer metrics reported in the main text.

\begin{table}[h]
\centering
\caption{
Additional biological-structure metrics on PBMC and BMMC.
Higher values indicate better performance for all metrics.
}
\label{tab:additional_bio_metrics}
\resizebox{0.85\textwidth}{!}{
\begin{tabular}{lcccccc}
\toprule
\multirow{2}{*}{\textbf{Method}}
& \multicolumn{3}{c}{\textbf{PBMC}}
& \multicolumn{3}{c}{\textbf{BMMC}} \\
\cmidrule(lr){2-4}
\cmidrule(lr){5-7}
& \textbf{Balanced LTA} $\uparrow$
& \textbf{Macro-F1} $\uparrow$
& \textbf{mAP} $\uparrow$
& \textbf{Balanced LTA} $\uparrow$
& \textbf{Macro-F1} $\uparrow$
& \textbf{mAP} $\uparrow$ \\
\midrule
Seurat v5
& 0.764 & 0.762 & 0.809
& 0.648 & 0.652 & 0.642 \\

GLUE
& 0.877 & 0.876 & 0.823
& 0.712 & 0.714 & 0.651 \\

MultiVI
& 0.721 & 0.698 & 0.682
& 0.632 & 0.628 & 0.483 \\

Cobolt
& 0.768 & 0.769 & 0.521
& 0.589 & 0.584 & 0.312 \\

FlowAlign
& 0.894 & 0.896 & 0.787
& 0.765 & 0.770 & 0.576 \\

\bottomrule
\end{tabular}
}
\end{table}

\subsection{Seed Variability of Main Single-Cell Benchmarks}
\label{app:seed_variance}

We report mean \(\pm\) standard deviation over three independent runs with a fixed train/validation/test split. 
For stochastic neural methods, the runs vary random initialization and training randomness. 
For deterministic pipelines such as Seurat v5, repeated runs under the same configuration produce negligible variation so all standard deviations are 0.000.

\begin{table}[H]
\centering
\caption{
Seed variability on the PBMC benchmark.
}
\label{tab:seed_variance_pbmc}
\small
\resizebox{\textwidth}{!}{
\begin{tabular}{lcccccc}
\toprule
\textbf{Method}
& \textbf{FOSCTTM} $\downarrow$
& \textbf{Cell match} $\uparrow$
& \textbf{LTA} $\uparrow$
& \textbf{NMI} $\uparrow$
& \textbf{ARI} $\uparrow$
& \textbf{ASW-O} $\uparrow$ \\
\midrule
Seurat v5
& \(0.058 \pm 0.000\)
& \(0.019 \pm 0.000\)
& \(0.850 \pm 0.000\)
& \(0.783 \pm 0.000\)
& \(0.785 \pm 0.000\)
& \(0.995 \pm 0.000\) \\

GLUE
& \(0.044 \pm 0.003\)
& \(0.016 \pm 0.003\)
& \(0.914 \pm 0.019\)
& \(0.802 \pm 0.021\)
& \(0.813 \pm 0.018\)
& \(0.979 \pm 0.009\) \\

MultiVI
& \(0.093 \pm 0.011\)
& \(0.007 \pm 0.002\)
& \(0.834 \pm 0.049\)
& \(0.784 \pm 0.042\)
& \(0.753 \pm 0.051\)
& \(0.830 \pm 0.035\) \\

Cobolt
& \(0.088 \pm 0.014\)
& \(0.012 \pm 0.003\)
& \(0.812 \pm 0.034\)
& \(0.799 \pm 0.042\)
& \(0.786 \pm 0.050\)
& \(0.921 \pm 0.021\) \\

FlowAlign
& \(0.039 \pm 0.004\)
& \(0.141 \pm 0.009\)
& \(0.920 \pm 0.012\)
& \(0.872 \pm 0.020\)
& \(0.885 \pm 0.025\)
& \(0.978 \pm 0.006\) \\
\bottomrule
\end{tabular}
}
\end{table}

\begin{table}[H]
\centering
\caption{
Seed variability on the BMMC benchmark.
}
\label{tab:seed_variance_bmmc}
\small
\resizebox{\textwidth}{!}{
\begin{tabular}{lcccccc}
\toprule
\textbf{Method}
& \textbf{FOSCTTM} $\downarrow$
& \textbf{Cell match} $\uparrow$
& \textbf{LTA} $\uparrow$
& \textbf{NMI} $\uparrow$
& \textbf{ARI} $\uparrow$
& \textbf{ASW-O} $\uparrow$ \\
\midrule
Seurat v5
& \(0.045 \pm 0.000\)
& \(0.008 \pm 0.000\)
& \(0.756 \pm 0.000\)
& \(0.714 \pm 0.000\)
& \(0.655 \pm 0.000\)
& \(0.978 \pm 0.000\) \\

GLUE
& \(0.056 \pm 0.002\)
& \(0.015 \pm 0.002\)
& \(0.745 \pm 0.013\)
& \(0.701 \pm 0.017\)
& \(0.689 \pm 0.019\)
& \(0.981 \pm 0.003\) \\

MultiVI
& \(0.061 \pm 0.006\)
& \(0.011 \pm 0.003\)
& \(0.777 \pm 0.047\)
& \(0.642 \pm 0.059\)
& \(0.585 \pm 0.058\)
& \(0.782 \pm 0.010\) \\

Cobolt
& \(0.097 \pm 0.012\)
& \(0.006 \pm 0.002\)
& \(0.742 \pm 0.024\)
& \(0.740 \pm 0.032\)
& \(0.512 \pm 0.035\)
& \(0.946 \pm 0.006\) \\

FlowAlign
& \(0.019 \pm 0.001\)
& \(0.050 \pm 0.005\)
& \(0.836 \pm 0.014\)
& \(0.797 \pm 0.019\)
& \(0.768 \pm 0.021\)
& \(0.988 \pm 0.003\) \\
\bottomrule
\end{tabular}
}
\end{table}

\subsection{FlowAlign as a Plug-in Regularizer}
\label{app:peakvi_regularizer}
We further evaluate whether FlowAlign can be used as a plug-in regularizer for an existing ATAC representation model in a PBMC proof-of-concept experiment. 
We first train scVI on scRNA-seq data and PeakVI on scATAC-seq data, obtaining separate RNA and ATAC latent spaces. 
The scVI RNA latent space is kept fixed as the target space, and FlowAlign is used to reshape the pretrained PeakVI ATAC latent space toward this target. 
Starting from the pretrained PeakVI model, we fine-tune it with the original PeakVI peak reconstruction loss and the FlowAlign path-consistency loss. 
This experiment tests whether FlowAlign can improve alignment to the fixed RNA target space without substantially sacrificing the original peak-accessibility modeling ability of PeakVI.

We evaluate target-space alignment using FOSCTTM, Cell Match (R@1), LTA, and NMI. 
The pretrained PeakVI row measures the unregularized ATAC latent before alignment and serves as the initialization baseline for this regularization experiment. 
To measure whether the PeakVI decoder is preserved, we report peak reconstruction AUROC and AUPR, computed by comparing decoder-predicted accessibility probabilities with the binarized peak-accessibility matrix over held-out cells and evaluated peaks. 
Peak AUROC measures how well the decoder ranks truly accessible peaks above inaccessible peaks across classification thresholds. 
Peak AUPR summarizes the precision--recall trade-off for recovering accessible peaks and is especially informative for sparse peak-accessibility matrices.

As shown in Table~\ref{tab:peakvi_regularizer}, FlowAlign regularization substantially improves the alignment between the PeakVI ATAC latent space and the fixed scVI RNA latent space, while largely preserving peak reconstruction performance. 
Although this regularized setting does not match the standalone FlowAlign alignment performance reported in the main experiments, it reaches a range comparable to established multi-omics alignment baselines. 
Importantly, this improvement is obtained while retaining the original PeakVI decoder and reconstruction objective. 
These results suggest that FlowAlign can reshape a pretrained ATAC latent space toward a fixed RNA target space as a plug-in regularizer, improving target-space alignment without substantially sacrificing the task-specific modeling ability of the original ATAC model.
\begin{table}[H]
\centering
\caption{
Performance of FlowAlign as a plug-in regularizer for PeakVI-to-scVI latent alignment on PBMC.
FlowAlign improves alignment of the pretrained PeakVI ATAC latent space to the fixed scVI RNA space while largely preserving peak reconstruction performance.
}
\label{tab:peakvi_regularizer}
\resizebox{0.85\textwidth}{!}{
\begin{tabular}{lcccccc}
\toprule
\textbf{Model}
& \textbf{FOSCTTM} $\downarrow$
& \textbf{R@1} $\uparrow$
& \textbf{LTA} $\uparrow$
& \textbf{NMI} $\uparrow$
& \textbf{Peak AUROC} $\uparrow$
& \textbf{Peak AUPR} $\uparrow$ \\
\midrule
Pretrained PeakVI
& 0.548 & 0.000 & 0.047 & 0.457 & 0.782 & 0.637 \\
PeakVI + FlowAlign reg.
& 0.068 & 0.046 & 0.815 & 0.801 & 0.765 & 0.630 \\
\bottomrule
\end{tabular}
}
\end{table}

\subsection{Held-Out Stage-1 Field Diagnostics}
\label{app:field_diagnostics}

We evaluate the frozen Stage-1 velocity field on held-out data using 32 Monte Carlo samples per cell.
For each source condition, we compare the normalized velocity residual on the true paired path with that on a path using a randomly shuffled target endpoint.
On PBMC, the true and shuffled residuals are 0.03185 and 0.11345, respectively, yielding a ratio of 3.562 and a ROC-AUC of 0.925.
On BMMC, the corresponding residuals are 0.02302 and 0.07355, with a ratio of 3.195 and a ROC-AUC of 0.921.
Using 2,000 bootstrap resamples, the 95\% confidence intervals for the residual ratios are [3.414, 3.725] and [3.145, 3.244], and those for ROC-AUC are [0.914, 0.936] and [0.917, 0.926], respectively.
These results indicate that the learned field assigns lower residuals to correct paired paths than to mismatched endpoints on held-out data, supporting the empirical premise of the path-consistency criterion.
However, these diagnostics do not directly establish the local smoothness assumption used in Proposition~\ref{prop:endpoint_control}.

\newpage
\section*{NeurIPS Paper Checklist}

\begin{enumerate}

\item {\bf Claims}
    \item[] Question: Do the main claims made in the abstract and introduction accurately reflect the paper's contributions and scope?
    \item[] Answer: \answerYes{} 
    \item[] Justification: Each claim in the abstract and introduction is supported by the corresponding method, theory, and experimental sections of the paper.

\item {\bf Limitations}
    \item[] Question: Does the paper discuss the limitations of the work performed by the authors?
    \item[] Answer: \answerYes{} 
    \item[] Justification: We discuss the limitations in the Conclusion, Limitations, and Future Work section.

\item {\bf Theory assumptions and proofs}
    \item[] Question: For each theoretical result, does the paper provide the full set of assumptions and a complete (and correct) proof?
    \item[] Answer: \answerYes{} 
    \item[] Justification: Assumptions are in Method section and proofs are in appendix.

    \item {\bf Experimental result reproducibility}
    \item[] Question: Does the paper fully disclose all the information needed to reproduce the main experimental results of the paper to the extent that it affects the main claims and/or conclusions of the paper (regardless of whether the code and data are provided or not)?
    \item[] Answer: \answerYes{} 
    \item[] Justification: We have the experiment implementation details in appendix.

\item {\bf Open access to data and code}
    \item[] Question: Does the paper provide open access to the data and code, with sufficient instructions to faithfully reproduce the main experimental results, as described in supplemental material?
    \item[] Answer: \answerYes{} 
    \item[] Justification: The datasets used in this work are publicly available. Our implementation, experimental configurations, and reproduction instructions are available at \url{https://github.com/yrzzz/FlowAlign}.

\item {\bf Experimental setting/details}
    \item[] Question: Does the paper specify all the training and test details (e.g., data splits, hyperparameters, how they were chosen, type of optimizer) necessary to understand the results?
    \item[] Answer: \answerYes{} 
    \item[] Justification: We provide a specific appendix with the experimental details.

\item {\bf Experiment statistical significance}
    \item[] Question: Does the paper report error bars suitably and correctly defined or other appropriate information about the statistical significance of the experiments?
    \item[] Answer: \answerYes{} 
    \item[] Justification: We provided variability results in appendix.

\item {\bf Experiments compute resources}
    \item[] Question: For each experiment, does the paper provide sufficient information on the computer resources (type of compute workers, memory, time of execution) needed to reproduce the experiments?
    \item[] Answer: \answerYes{} 
    \item[] Justification: We specified what type of GPU we used.
    
\item {\bf Code of ethics}
    \item[] Question: Does the research conducted in the paper conform, in every respect, with the NeurIPS Code of Ethics \url{https://neurips.cc/public/EthicsGuidelines}?
    \item[] Answer: \answerYes{} 
    \item[] Justification: We reviewed the policy.

\item {\bf Broader impacts}
    \item[] Question: Does the paper discuss both potential positive societal impacts and negative societal impacts of the work performed?
    \item[] Answer: \answerYes{} 
    \item[] Justification: We discuss potential scientific benefits and limitations in the Introduction and Conclusion.
    
\item {\bf Safeguards}
    \item[] Question: Does the paper describe safeguards that have been put in place for responsible release of data or models that have a high risk for misuse (e.g., pre-trained language models, image generators, or scraped datasets)?
    \item[] Answer: \answerNA{} 
    \item[] Justification: The paper does not release high-risk models, scraped datasets, or generative systems with substantial misuse potential. The work focuses on representation alignment for synthetic and public single-cell multiome data.

\item {\bf Licenses for existing assets}
    \item[] Question: Are the creators or original owners of assets (e.g., code, data, models), used in the paper, properly credited and are the license and terms of use explicitly mentioned and properly respected?
    \item[] Answer: \answerYes{} 
    \item[] Justification: We use publicly available single-cell multiome datasets and open-source software packages

\item {\bf New assets}
    \item[] Question: Are new assets introduced in the paper well documented and is the documentation provided alongside the assets?
    \item[] Answer: \answerNA{} 
    \item[] Justification: The paper does not introduce a new dataset, benchmark, or pretrained model asset.

\item {\bf Crowdsourcing and research with human subjects}
    \item[] Question: For crowdsourcing experiments and research with human subjects, does the paper include the full text of instructions given to participants and screenshots, if applicable, as well as details about compensation (if any)? 
    \item[] Answer: \answerNA{} 
    \item[] Justification: The paper does not involve crowdsourcing experiments

\item {\bf Institutional review board (IRB) approvals or equivalent for research with human subjects}
    \item[] Question: Does the paper describe potential risks incurred by study participants, whether such risks were disclosed to the subjects, and whether Institutional Review Board (IRB) approvals (or an equivalent approval/review based on the requirements of your country or institution) were obtained?
    \item[] Answer: \answerNA{} 
    \item[] Justification: The paper does not involve new data collection from human subjects or interaction with study participants.

\item {\bf Declaration of LLM usage}
    \item[] Question: Does the paper describe the usage of LLMs if it is an important, original, or non-standard component of the core methods in this research? Note that if the LLM is used only for writing, editing, or formatting purposes and does \emph{not} impact the core methodology, scientific rigor, or originality of the research, declaration is not required.
    \item[] Answer: \answerNo{} 
    \item[] Justification: LLMs were only used for grammatical purposes.

\end{enumerate}

\end{document}